\documentclass[letterpaper, 10 pt,conference]{ieeeconf}  
\IEEEoverridecommandlockouts                              
\usepackage[utf8]{inputenc}
\usepackage[english]{babel}
\usepackage{times}

\usepackage{amssymb,amsmath,amsthm}
\usepackage{cite}
\usepackage{multirow}
\usepackage{graphicx}
\usepackage[table,xcdraw]{xcolor}
\usepackage{psfrag}
\usepackage{color}
\usepackage[us]{datetime}
\usepackage{algorithm}
\usepackage{algpseudocode}
\usepackage{balance}
\usepackage{makecell}
\usepackage{mathtools}
\usepackage{array}
\usepackage{lipsum}
\usepackage{float}
\usepackage{color}
\usepackage{hyperref}
\usepackage{cleveref}
\usepackage{siunitx}
\usepackage[export]{adjustbox}
\usepackage{subcaption}
\usepackage{textcomp}

\renewcommand{\baselinestretch}{1}
\usepackage{amsmath}

\newcommand{\vect}[1]{\boldsymbol{#1}}
\newcommand{\mat}[1]{\boldsymbol{#1}}

\newcommand{\diffs}[3]{\frac{\partial^2 #1}{
\ifx#2#3 
\partial #2^2
\else
\partial #2 \partial #3
\fi
}}

\newcommand{\av}{\vect{a}}

\newcommand{\fv}{\vect{f}}

\newcommand{\pv}{\vect{p}}
\newcommand{\dpv}{\dot{\vect{p}}}
\newcommand{\ddpv}{\ddot{\vect{p}}}
\newcommand{\qv}{{\vect{q}}}
\newcommand{\dqv}{\dot{\vect{q}}}

\newcommand{\xv}{\vect{x}}
\newcommand{\dxv}{\dot{\vect{x}}}

\newcommand{\IIm}{\mat{I}}

\newcommand{\Am}{\mat{A}}
\newcommand{\Bm}{\mat{B}}

\newcommand{\Jm}{\mat{J}}

\newcommand{\Km}{\mat{K}}

\newcommand{\Pm}{\mat{P}}

\newcommand{\Vm}{\mat{V}}

\newcommand{\Lambdam}{\mat{\Lambda}}

\usepackage{bm}

\graphicspath{{figures/}}
\usepackage{tabularx,booktabs}
\usepackage{caption}
\makeatletter
\algrenewcommand\ALG@beginalgorithmic{\footnotesize}
\makeatother

\def\BibTeX{{\rm B\kern-.05em{\sc i\kern-.025em b}\kern-.08em
    T\kern-.1667em\lower.7ex\hbox{E}\kern-.125emX}}

\title{\LARGE \bf Kinematic Control of Redundant Robots with Online \\ Handling of Variable Generalized Hard Constraints}

\author{Amirhossein Kazemipour$^\ast$, Maram Khatib$^\ast$, Khaled Al Khudir$^{\ast \ast}$, Claudio Gaz$^\ast$, Alessandro De Luca$^\ast$
\thanks{$^\ast$Dipartimento di Ingegneria Informatica, Automatica e Gestionale, Sapienza Universit\`a di Roma, Via Ariosto 25, 00185 Roma, Italy. Emails: \href{mailto:amrkzp@gmail.com}{amrkzp@gmail.com}, \{\href{mailto:khatib@diag.uniroma1.it}{khatib}, \href{mailto:gaz@diag.uniroma1.it}{gaz}, \href{mailto:deluca@diag.uniroma1.it}{deluca}\}@diag.uniroma1.it}
\thanks{$^{\ast \ast}$School of Mechanical, Aerospace
and Automotive Engineering, Coventry University, CV1 5FB Coventry, UK. Email: \href{mailto:khaled.alkhudir@coventry.ac.uk}{khaled.alkhudir@coventry.ac.uk}}}

\begin{document}
\maketitle

\begin{abstract}
We present a generalized version of the Saturation in the Null Space (SNS) algorithm of~\cite{flacco2015control} for the task control of redundant robots when hard inequality constraints are simultaneously present both in the joint and in the Cartesian space. These hard bounds should never be violated, are treated equally and in a unified way by the algorithm, and may also be varied, inserted or deleted online. 
When a joint/Cartesian bound saturates, the robot redundancy is exploited to continue fulfilling the primary task.
If no feasible solution exists, an optimal scaling procedure is applied to enforce directional consistency with the original task. 
Simulation and experimental results on different robotic systems demonstrate the efficiency of the approach. 
\end{abstract}

\section{Introduction}

A robot manipulator is redundant with respect to a given task when the number of its joints is larger than that strictly needed to perform the task. The additional degrees of freedom allow for a greater flexibility in the execution of the primary task. We usually take advantage of such redundancy for achieving secondary goals, such as avoiding collisions with workspace obstacles, maximize manipulability, stay away from kinematic singularities, or minimize energy consumption~\cite{chiaverini2016redundant}. The presence of joint and Cartesian inequality constraints is a critical issue in redundancy resolution. Robots should comply with hard constraints on position,  velocity and acceleration in their joint motion, typically coming from actuator limitations. Inequality constraints on the Cartesian motion may be present because of the nature of the task, or sometimes suddenly appear due to the unstructured environment in which robots operate. 

There are many ways to handle joint and Cartesian constraints in  kinematic control of robots. Classical methods use artificial potentials~\cite{khatib1986real}, with a number of control points chosen along the kinematic chain being pushed away from the critical boundaries~\cite{khatib2020task} and the associated control action taking place in the null space of the Jacobian of the primary task.  This method is simple and effective, but highly parameter-dependent. Moreover, oscillatory behaviors may arise when activating/deactivating the evasive maneuvers in the proximity of the constraints~\cite{khatib2020Irim}. In order to mitigate this undesired effect, the null-space projection term or the activation function may be designed in an incremental way~\cite{mansard2009unified,simetti2016novel}. Nonetheless, the selection of suitable gains is still required. Furthermore, when multiple tasks are present, incorporating the avoidance scheme into the original Stack of Tasks (SoT) will give to each inequality constraint a different  priority~\cite{mansard2009unified,simetti2016novel,sentis2005synthesis,sentis2006whole}. A framework dealing with adaptable Cartesian constraints with task scaling has been recently presented in~\cite{magrini2020}. However, this method is not able to manage a SoT with different priorities.

In order to deal with joint positional constraints, a common approach is to transform hard joint bounds into soft constraints, by adopting a suitable cost function whose minimization will keep the joint motions close to the center of their admissible ranges~\cite{liegeois1977automatic}. Alternatively, a weighted pseudo-inverse technique can be used~\cite{chan1995weighted}, which further penalizes joint motions when they approach their limits. These techniques, however, do not guarantee that the hard inequality constraints will be always satisfied, and so they may result in unfeasible solutions. Recently, constrained optimization has been applied to the inverse kinematics of redundant robots, transforming the given tasks into a Least Squares (LS) problem and looking for solutions within a feasible convex set. A general formulation within this paradigm, which extends the priority framework also to inequality tasks, has been presented in~\cite{kanoun2011kinematic}. 

\begin{figure}[!t]
\centering
\includegraphics[scale=0.35]{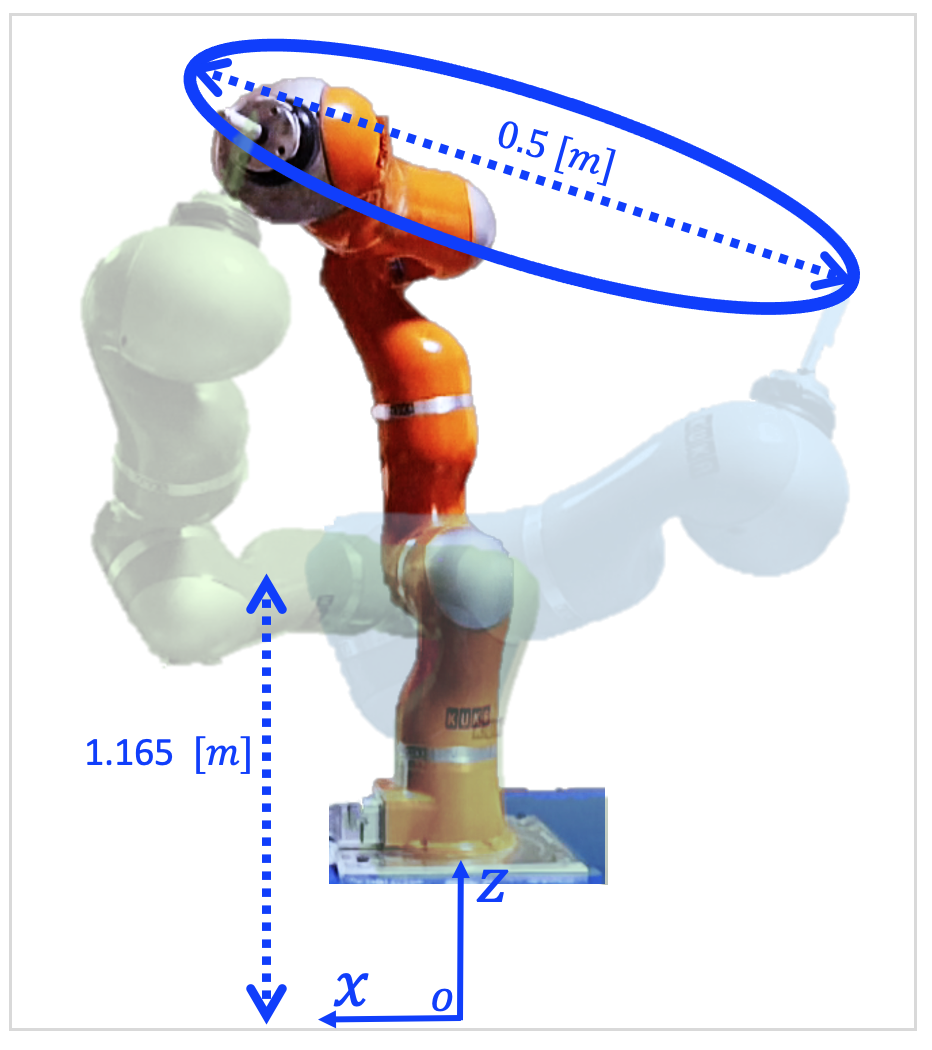}
\caption{The KUKA LWR IV robot used for experimental evaluation. The world frame is placed on the lab floor. The desired end-effector task, the initial (solid orange), intermediate (shaded blue), and final (shaded orange) robot configurations of the first experiment are shown.}
\label{fig:KUKA}
\vspace{-15pt}
\end{figure}

The LS formulation has the advantage of explicitly including hard bounds into a numerically solvable Quadratic Programming (QP) problem. It allows to incorporate both joint and Cartesian motion limits as inequality constraints~\cite{escande2014hierarchical,hoffman2018multi}. However, these numerical approaches are computationally slower than analytical solutions~\cite{flacco2013fast}. Moreover, the feasibility of the task cannot be enforced, and the solution will be realizable only if the original task is compatible with the set of inequality constraints. Otherwise, the relaxation of these constraints in a least square sense leads to a physical violation of the hard limits.

The Saturation in the Null Space (SNS) algorithm introduced in~\cite{flacco2015control} is capable of resolving part of these issues, by linking QP to the SoT approach. In the original paper, the constraints on joint motion are regarded as hard bounds (i.e., they cannot be relaxed in a least-square sense) and treated out of the SoT. So far, Cartesian bounds have not been treated explicitly, but rather approximated in the joint space as soft constraints. Accordingly, there is no guarantee that the robot will strictly comply with the hard Cartesian constraints.

On the other hand, in~\cite{osorio2019physical}, both joint and Cartesian inequality constraints have been taken explicitly into account in the SoT for torque-controlled manipulators. In this approach, hard joint limits should always be given the highest priority over all other constraints. However, when both Cartesian and joint constraints are violated, the algorithm becomes unreliable because other lower priority limits will be treated as soft constraints.
Therefore, when the primary task cannot be realized due to inequality constraints, even the directional component of the task will no longer be guaranteed, leading to a global deformation in its execution.
Furthermore, at each iteration, the algorithm in~\cite{osorio2019physical} sets all the joint/Cartesian commands that exceed their limits to their saturation level. This choice is neither necessary nor optimal, and it often results in high-frequency oscillations over time~\cite{flacco2015control}. Saturating only the most critical command at a time allows instead other violated constraints to be possibly recovered in subsequent iterations of the algorithm~\cite{flacco2012prioritized}.

Building on our preliminary results in~\cite{kazemipour2021Irim}, we generalize the original SNS algorithm in~\cite{flacco2015control} with the following contributions.
\begin{itemize}
    \item A single augmented vector is defined that considers all joint and Cartesian inequality constraints explicitly. This vector can be adapted online, without any parameter tuning phase, to follow any desired modification (addition or removal) in the set of task constraints. 
    \item In the proposed algorithm, presented here at the velocity command level, all joint/Cartesian inequality constraints are treated equally. Accordingly, the hard bounds are always respected strictly. This is independent of the primary task (or of the SoT and its related priority management, when considering multiple equality tasks).
    \item The primary task $\dxv$ is relaxed optimally by keeping its geometric direction, if and only if no feasible solution exists. Differently from~\cite{flacco2015control}, if $\dxv$ exceeds any Cartesian constraint, it is saturated to its associated limit. 
    \item The algorithm applies the saturation technique in both the joint and the Cartesian space. Again, this is unlike~\cite{flacco2015control}, where saturation is only applied in the joint space.   
\end{itemize}
The resulting control algorithm can be viewed as a general tool that can be easily used in any robot application, such as human-robot collaboration tasks~\cite{khatib2021human}. Its main feature is in fact an overall efficiency and the adaptability to time-varying hard constraints that may be generated or deleted online based on sensor information. The validation of the basic algorithm is carried out with a simulation and two different experiments that are illustrated also in the accompanying video. 

The rest of the paper is organized as follows. Section~\ref{sec:preliminaries} introduces the framework for incorporating generalized constraints. In Sec.~\ref{sec:modifiedSNS}, the new kinematic control algorithm is presented at the velocity command level. Simulation results on a planar 6R manipulator and experiments on the 7R KUKA LWR IV robot results are reported and discussed in Sec.~\ref{sec:experiments}. Conclusions are summarized in Sec.~\ref{sec:conclusion}.

\section{Generalized Hard Constraints}
\label{sec:preliminaries}

Consider a robot manipulator with $n$ joints and a single $m$-dimensional task, with $m < n$, to be performed by its end effector (EE) and defined by
\begin{equation}
\xv=\fv(\qv), \qquad \Jm(\qv) = \frac{\partial\fv(\qv)}{\partial\qv}, 
\label{eq:task}
\end{equation}
where $\qv \in \mathbb{R}^n$ is the joint position vector and the ${m \times n}$ task Jacobian matrix $\Jm$ has less rows than columns.
Assuming that the robot is commanded by a kinematic control law at the velocity level, we solve the inverse differential kinematics as
\begin{equation}
\dqv=\Jm^{\#}(\qv) \dxv,
\label{eq:first_diff}
\end{equation}
where $\Jm^{\#}$ is the Moore–Penrose pseudoinverse of $\Jm$.
The command~(\ref{eq:first_diff}) is the minimum norm joint velocity corresponding to the desired task velocity $\dxv$. It is the preferred solution in the absence of the constraints, either in the joint or in the Cartesian space, that we shall consider next. 

Define the position, velocity, and acceleration limits of each joint, $j=1,\dots,n$, respectively as 
\begin{equation}
\begin{array}{c}
Q_j^{min} \leq q_j \leq Q_j^{max}, \\[6pt]
V_j^{min} \leq \dot{q}_j \leq V_j^{max},\\[6pt]
\Lambda_j^{min} \leq \ddot{q}_j \leq \Lambda_j^{max}.
\end{array}
\label{eq.4.1j}
\end{equation}
Accordingly, the box constraints for each joint can be defined, at the velocity level, as
\begin{equation}
\begin{array}{l}
\dot{Q}_{min,j}=\\[4pt]
\displaystyle \max \left\{ \frac{Q_j^{min}-q_j}{T},V_j^{min},-\sqrt{2 \Lambda_j^{max} \left( q_j-Q_j^{min}\right)} \right\}, \\[12pt]
\dot{Q}_{max,j}=\\[4pt]
\displaystyle \min \left\{ \frac{Q_j^{max}-q_j}{T},V_j^{max},\sqrt{2 \Lambda_j^{max} \left( Q_j^{max}-q_j\right)} \right\},
\end{array}
\label{eq:cons1}
\end{equation}
where $T$ is the sampling time. Consider next $r$ generic Cartesian control points distributed along the robot body, each of dimension $d_i \in \{1,2,3\}$, $i=1,\dots, r$. The desired position, velocity, and acceleration limits for each control point $i$ can be defined as
\begin{equation}
\begin{array}{c}
\Pm_{cp,i}^{min} \leq \pv_{cp,i} \leq \Pm_{cp,i}^{max}, \\[6pt]
\Vm_{cp,i}^{min} \leq \dpv_{cp,i} \leq \Vm_{cp,i}^{max},\\[6pt]
\Lambdam_{cp,i}^{min} \leq \ddpv_{cp,i} \leq \Lambdam_{cp,i}^{max},
\end{array}
\label{eq.4.1}
\end{equation}
where $\pv_{cp,i}\in \mathbb{R}^{d_i}$ is the position of the $i$-th control point. As before, the box of constraints for each control point can be defined, at the velocity level, as
\begin{equation}
\begin{array}{l}
\dot{\Pm}_{cp,i}^{min}=\\[4pt]
\displaystyle \max \left\{ \frac{\Pm_{cp,i}^{min}-\pv_{cp,i}}{T},\Vm_{cp,i}^{min}, -\sqrt{2 \Lambdam_{cp,i}^{max} \left( \pv_{cp,i}-\Pm_{cp,i}^{min}\right)} \right\}, \\[12pt]
\dot{\Pm}_{cp,i}^{max}=\\[4pt]
\displaystyle \min \left\{ \frac{\Pm_{cp,i}^{max}-\pv_{cp,i}}{T},\Vm_{cp,i}^{max}, \sqrt{2 \Lambdam_{cp,i}^{max} \left( \Pm_{cp,i}^{max}-\pv_{cp,i}\right)} \right\},
\end{array}
\label{eq:cons2}
\end{equation}
 
To take into account all the inequality constraints in~(\ref{eq.4.1j}) and~(\ref{eq.4.1}) while executing the desired task $\dxv$, we define the augmented vector
\begin{equation}
\av = \left(\!\begin{array}{ccccc}
\qv^T & \pv_{cp,1}^T & \pv_{cp,2}^T & \dots & \pv_{cp,r}^T\end{array}\!\right)^T,
\label{eq:a}
\end{equation}
and the augmented matrix 
\begin{equation}
\Am = \left(\!\begin{array}{ccccc}
		\IIm & \Jm_{cp,1}^T & \Jm_{cp,2}^T & \dots & \Jm_{cp,r}^T
\end{array}\!\right)^T,
\label{eq:augment_mat}
\end{equation}
where $\IIm$ is the $(n \times n)$ identity matrix and $\Jm_{cp,i}$ is the $(d_i \times n)$ Jacobian of the $i$-th control point position. Accordingly, at a generic time instant $t=t_k=kT$, it is possible to define the generalized inequality constraints at the velocity level as
\begin{equation}
\Bm_{min} (t_k) \leq \dot{\av}(\qv,\dqv) \leq \Bm_{max} (t_k),
\label{eq:box}
\end{equation}
where $\Bm_{min}$ and $\Bm_{max}$ are the general limits augmented matrices and defined as
\begin{equation}
\begin{array}{rcl}
\Bm_{min}  &\!\!\!\!=\!\!\!\!& \left(\!\begin{array}{cccccc}
\dot{Q}_{min,1} & \!\!\!\!\dots\!\!\!\! & \dot{Q}_{min,n} &\!
\dot{\Pm}_{cp,1}^{min^T} & \!\!\!\!\dots\!\!\!\! & \dot{\Pm}_{cp,r}^{min^T}
\end{array}\!\right)^T\!, \\[4pt]
\Bm_{max}  &\!\!\!\!=\!\!\!\!& \left(\!\begin{array}{cccccc}
\dot{Q}_{max,1} & \!\!\!\!\dots\!\!\!\! & \dot{Q}_{max,n} &\!
\dot{\Pm}_{cp,1}^{max^T} & \!\!\!\!\dots\!\!\!\! & \dot{\Pm}_{cp,r}^{max^T}
\end{array}\!\right)^T\!.
\end{array}
\label{eq:Gbound}
\end{equation}

Satisfying the generalized box of constraints in~(\ref{eq:Gbound}) leads to impose strictly the original position and velocity bounds, by considering also the maximum acceleration limits. Note that, when the robot control law is defined at the velocity level, the acceleration limits can be treated only as soft constraints.

\section{The Generalized SNS Algorithm}
\label{sec:modifiedSNS}
\begin{algorithm}[t]
	\caption{SNS with generalized inequality constraints}
	\begin{algorithmic}[2]
		\State$\dqv_N \gets \mathbf{0},\; s^{\ast} \gets 0,\; \Pm \gets \IIm, \; \Am_{lim} \gets \text{null},\; \dot{\av}_N \gets \text{null}$
		\Repeat
		\State $\text{limits\_violated} \gets \text{FALSE}$
		
		\State $ \dqv \gets \dqv_N + \left( \Jm\,\Pm \right)^{\#}  \left( \dxv - \Jm\,\dqv_N \right)$
		\State $ \dot{\av} \gets \Am\,\dqv $
		\If{$ \exists \,h \in \, \left[1:n+\Sigma_1^r d_i\right]: \,  (\dot{a}_h < {{b}}_{min,h}) \, \lor \,  (\dot{a}_h > {{b}}_{max,h})  $}
		\State $\text{limits\_violated} \gets \text{TRUE}$
		\State $ \boldsymbol{\alpha} \gets \Am \left( \Jm\,\Pm \right)^{\#}\dxv$
		\State $ \boldsymbol{\beta} \gets \dot{\av} - \boldsymbol{\alpha}$
		\State $s_k \gets \text{getTaskScalingFactor}(\boldsymbol{\alpha},\boldsymbol{\beta}) $
		\State $ k \gets \{\text{the most critical constraint}\} $
		\If{$ s_k > s^\ast$}
		\State $ s^\ast \gets  s_k $
		\State $ \dqv_N^\ast \gets \dqv_N,\;\Pm^\ast \gets  \Pm$
		\EndIf
		\State $ \Am_{lim} \gets \text{concatenate}(\Am_{lim},\Am_k) $
		\State $ \dot{a}_{N} \gets \begin{cases}
			\text{concatenate}(\dot{a}_{N},{{b}}_{max,k})\;\;\;\;\;\; \text{if}\;\;(\dot{a}_h > {{b}}_{max,k})\\
			\text{concatenate}(\dot{a}_{N},{{b}}_{min,k})\;\;\;\;\;\; \text{if}\;\;(\dot{a}_h < {{b}}_{min,k})\\\end{cases} $
		\State $ \Pm \gets  \IIm - \left( \Am_{lim} \right)^{\#}\left(\Am_{lim}\right)$
		\If{$ \text{rank}(\Jm \Pm) < m \land k \not\in \{\text{primary task}\}$}
		\State $ \dqv \gets \dqv_N^\ast + \left( \Jm\,\Pm^\ast \right)^{\#}  \left( s^\ast\dxv - \Jm\,\dqv_N^\ast \right)  $
		\State $\text{limits\_violated} \gets \text{FALSE} $ 
		\EndIf
		\EndIf
		\State $ \dqv_N \gets \left( \Am_{lim} \right)^{\#} \, \dot{a}_{N} $
		\Until{$\text{limits\_violated} = \text{TRUE}$}
		\State $ \dqv_{SNS} \gets \dqv$
	\end{algorithmic}
\label{basic:vel:single}
\end{algorithm}

\begin{algorithm}[t]
	\caption{Optimal task scaling factor}
	\begin{algorithmic}[2]
    \Function{getTaskScalingFactor}{$\boldsymbol{\alpha},\boldsymbol{\beta}$}
        \For {$h \leftarrow 1: n+\Sigma_1^r d_i$}
            \State $L_h \gets b_{min,h}-\beta_h$
            \State $U_h \gets b_{max,h}-\beta_h$
                \If{$ {\alpha}_{h} < 0 \, \land \,  L_h < 0 $}
                    \If{$ \alpha_h < L_h $}
                    \State $s_h \gets L_h^{}/\alpha_h$
                    \Else
                    \State $s_h \gets 1 $
                    \EndIf
                \ElsIf{$\alpha_h>0 \, \land \, U_h > 0$}
                    \If{$ \alpha_h > U_h $}
                    \State $s_h \gets U_h^{}/\alpha_h$
                    \Else
                    \State $s_h \gets 1 $
                    \EndIf
                \Else
                    \State $s_h \gets 0$
                \EndIf
        \EndFor
    \State $s \gets \min s_h$\\
    \Return $s$
    \EndFunction
	\end{algorithmic}
\label{scaling_factor}
\end{algorithm}

We have revisited the original SNS algorithm in~\cite{flacco2015control} so as to cover also the generalized constraints~(\ref{eq:box}). We highlight here the main introduced differences.

The pseudo-code of the proposed scheme is presented as \Cref{basic:vel:single}. The method starts by initializing: a projection matrix $\Pm = \IIm$, a null-space joint velocity vector $\dqv_N = \bm{0}$, a scaling factor $s^\ast=0$, a null-space augmented velocity vector $\dot{\av}_N=\text{null}$, and an augmented saturation matrix $\Am_{lim}=\text{null}$. 

\begin{figure}[!t]
\centering
\includegraphics[width=\columnwidth]{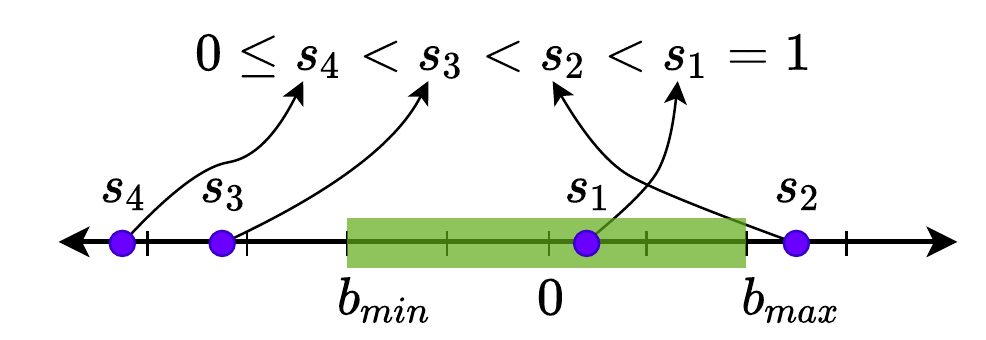}
\caption{The task scaling factor associated with each constraint is computed in~\Cref{scaling_factor}. The factor is maximum (equal to 1 for the original task) when the corresponding velocity falls within the admissible interval, i.e., $b_{min,h} \leq \dot{a}_h \leq b_{max,h}$. In all other cases, the scaling factor is less than 1, and the constraint becomes more critical as the associated velocity moves further away from the boundaries of the interval.}
\label{fig:constraints}
\end{figure}

On the basis of the minimum norm velocity solution, the current commanded joint velocity is given by
\begin{equation}\label{eq:proj_eq}
\dqv = \dqv_N + \left( \Jm\,\Pm \right)^{\#}  \left( \dxv - \Jm\,\dqv_N \right),
\end{equation}
which attempts to execute the desired task as efficiently as possible (with the lowest possible velocity norm) by enforcing the velocity of some overdriven joint/Cartesian constraints to saturation, thereby keeping the entire augmented velocity $\dot{\av}$ in~(\ref{eq:box}) within the desired constraints box.
If the solution in~(\ref{eq:proj_eq}) is acceptable under the constraints~(\ref{eq:box}), the algorithm terminates and outputs this velocity for controlling the robot at the current time instant $t_k$. On the other hand, if it violates one or more of the hard generalized constraints, the algorithm repeats the loop until an admissible solution is found.
This is accomplished by first calling~\Cref{scaling_factor} to determine the most critical constraint $k$, which corresponds to a constraint that has the smallest scaling factor $s_k$, among all limits (see \Cref{fig:constraints}). Next, only the $k$-th joint/Cartesian constraint is saturated to its limit during each iteration, and $\dot{a}_{N}$ and $\Am_{lim}$ are updated accordingly. Then, the solution~(\ref{eq:proj_eq}) is recomputed, and the procedure is repeated.

At line 18, the projection matrix $\bm{P}$ is constructed according to the current saturated constraints as
\begin{equation}
\Pm =  \IIm - \left( \Am_{lim} \right)^{\#}\left(\Am_{lim}\right),
\end{equation}
where $\Am_{lim}$ incorporates the generalized active constraints. At the current iteration, if the most critical constraint is associated with the $k$-th joint, then the $k$-th row of the identity matrix $\bm{I}_{n \times n}$ is extracted and augmented to $\Am_{lim}$. For the Cartesian constraints, a similar procedure is followed. For instance, if the most critical constraint is associated with the $x$-direction of the $k$-th control point, then the corresponding row of its Jacobian matrix $\bm{J}_{cp,k}^x$ is taken out and concatenated to $\Am_{lim}$.

At the end of each iteration, the algorithm checks if the robot is still redundant in executing the primary task under the currently active set of constraints ($\text{rank}(\Jm \Pm) \geq m$). This check fails when the task redundancy of the robot is exhausted, and there is no way to perform $\dxv$ under the given  constraints. In this case, the solution with the highest task scaling factor obtained so far (the value $s_k$ that is closest to 1) is applied
\begin{equation}
    \dqv = \dqv_N^\ast + \left( \Jm\,\Pm^\ast \right)^{\#}  \left( s^\ast\dxv - \Jm\,\dqv_N^\ast \right).
\end{equation}
This choice preserves the geometry of the task, although it scales down the intensity of the task velocity by the (optimal) factor $s^\ast$.

Note that, if the current most critical constraint is associated with $\dxv$, i.e., $k \in \{\text{primary task}\}$, e.g., when a control point coincides with the end-effector, then there is no need to scale down the task since its violated part is saturated to its limit, i.e., by modifying $\dot{a}_{N}$ and $\Am_{lim}$ accordingly. In this way, differently from~\cite{flacco2015control}, \Cref{basic:vel:single} is able to saturate the constraints in both the joint and the Cartesian space. Finally, we remark here that the proposed algorithm can be used to manage easily multiple Cartesian tasks with equal priority, e.g., for collision avoidance purposes.

\section{Numerical Results}
\label{sec:experiments}
The efficiency of the new algorithm has been evaluated in simulation, using a 6R planar manipulator ($n=6$), and with experiments on a KUKA LWR IV robot ($n=7$), see Figs.~\ref{fig:KUKA} and \ref{fig:motion}. For the presented case studies, a stabilizing feedback action is integrated into the desired EE velocity $\dxv$ of the primary task to compensate for any numerical errors as
\begin{equation}
\dxv = \dxv_d + \Km_p(\xv_d - \fv_{ee}(\qv)),
\end{equation}
where $\xv_d(\sigma)$ is the desired parametrized Cartesian path, $\sigma(t)$ is the timing law of the path parameter, $\fv_{ee}(\qv)$ is the robot direct kinematics, and $\Km_p>0$ is the (diagonal) control gain matrix of dimension $m$. A suitable rest-to-rest timing law is considered in each case. The actual motions of the robots are shown in the accompanying video.  

\subsection{Simulation with a 6R planar manipulator}

\begin{figure}[thpb]
	\centering
	\includegraphics[scale=0.55]{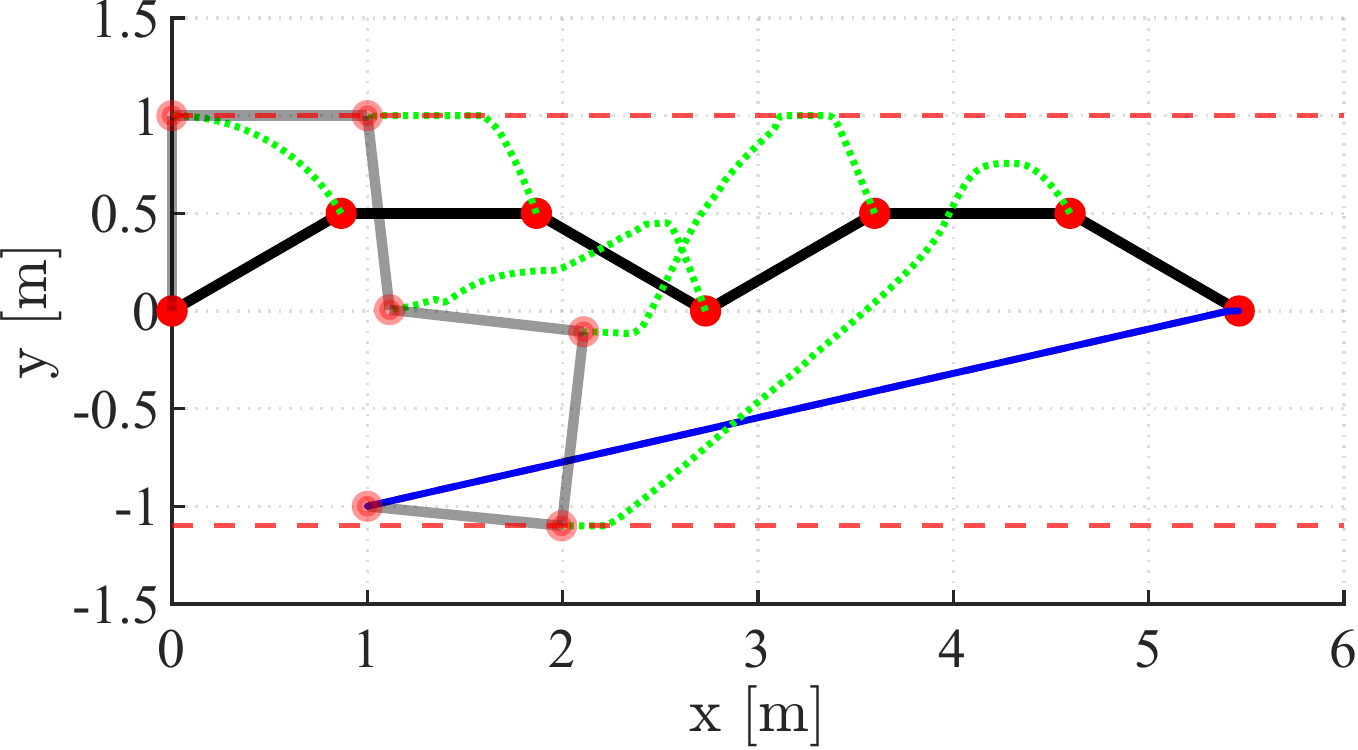}  
	\caption{Simulation. The 6R planar arm is shown in its initial (black) and final (gray) configurations. The robot joints (and the end effector) are represented by red circles. The desired end-effector path is the blue line, to be traced from right to left. The Cartesian position bounds are indicated by the two dashed red lines. The dotted green lines are the paths of the chosen control points during task execution.}
	\label{fig:motion}
\end{figure}

A verification of \Cref{basic:vel:single} has been done first through a MATLAB simulation. The EE of a 6R planar manipulator should  track a 2D linear path ($m=2$) in $T_{interval}=10$ [s] with a $5^{th}$-order polynomial timing law, see Fig.~\ref{fig:motion}. The control gain matrix is set to $\Km_p = \mbox{diag}\{2, 2\}$ and the  sampling time is $T = 1$~[ms].
The initial robot configuration is chosen as 
\begin{equation}
\qv_0 = \left(\!\begin{array}{cccccc} 30 & -30 & -30 &60 &-30 & -30 \end{array}\!\right)^T \ [\deg].
\end{equation}

In this example, the joint position and velocity limits in~(\ref{eq.4.1j}) are equal and symmetric for all joints $j=1,\dots,6$, where
\begin{equation}
\begin{split}
{Q}_j^{max} &= -{Q}_j^{min} = 90 \ \ [\deg], \\
{V}_j^{max} &= -{V}_j^{min} = \frac{90}{\pi} \ \ [\si{\deg/\second}]. 
\end{split}
\end{equation}
As for Cartesian constraints, we considered $r=5$ control points (each with $d_i=1$) along the robot body, located at the joints $j=2,\dots,6$. The Cartesian position and velocity limits in~(\ref{eq.4.1j}) are the same for all control points, and are imposed only along the $y$-direction:
\begin{equation}
\begin{array}{l}
P_{cp,1}^{max,y} = 1,  \ \ \ P_{cp,1}^{min,y} = -1.1 \ \mbox{[m]}, 
\\[8pt]
V_{cp,1}^{max,y} = 0.5, \ V_{cp,1}^{min,y} = -0.5 \ \mbox{[m/s]}.
\end{array}
\end{equation}

\begin{figure}[thpb]
	\centering
	\includegraphics[scale=0.4]{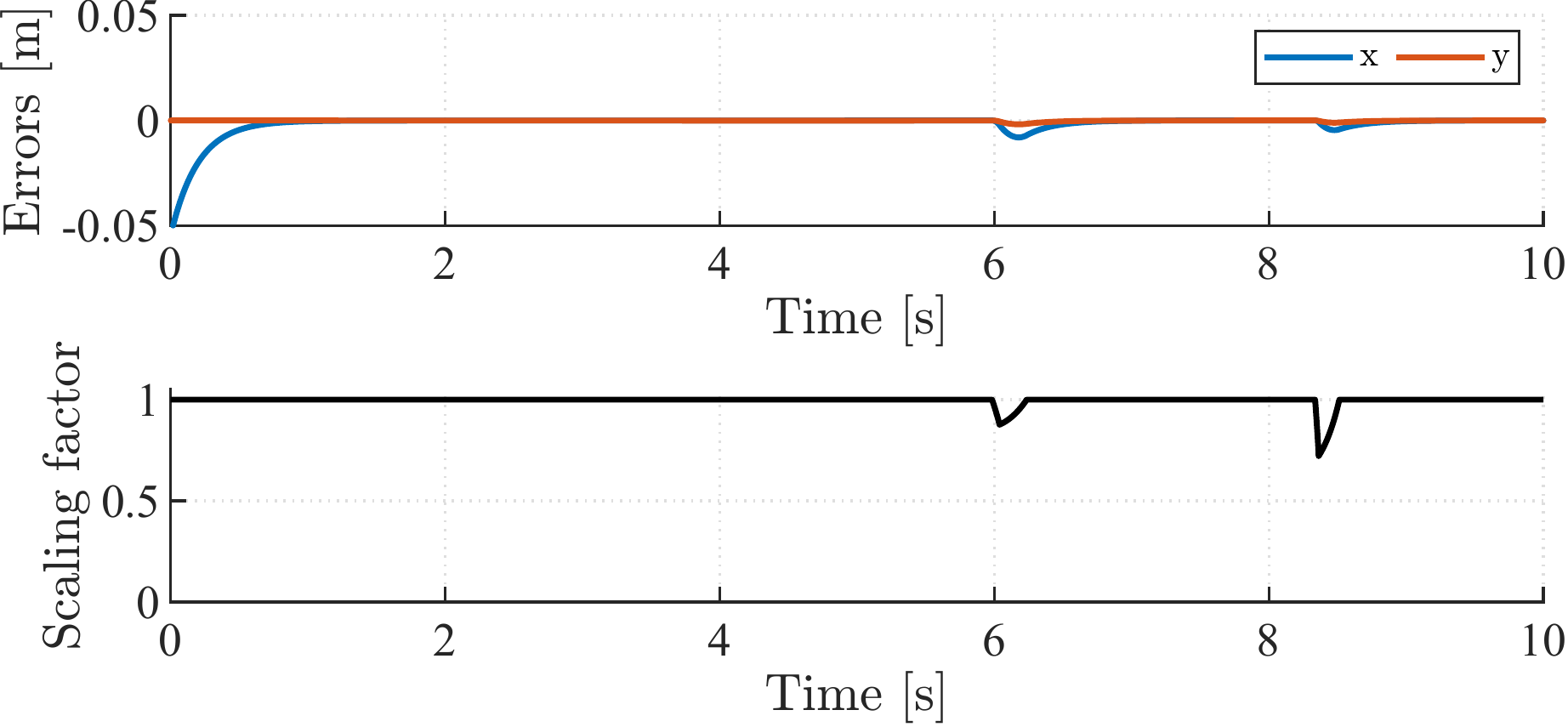}
	\caption{Simulation. The end-effector $x$ and $y$ position errors and the associated task scaling factor.}
	\label{fig:error}
\end{figure}

\begin{figure}[thpb]	
	\centering
	\begin{subfigure}{0.99\columnwidth}
		\centering
		\includegraphics[scale=0.45]{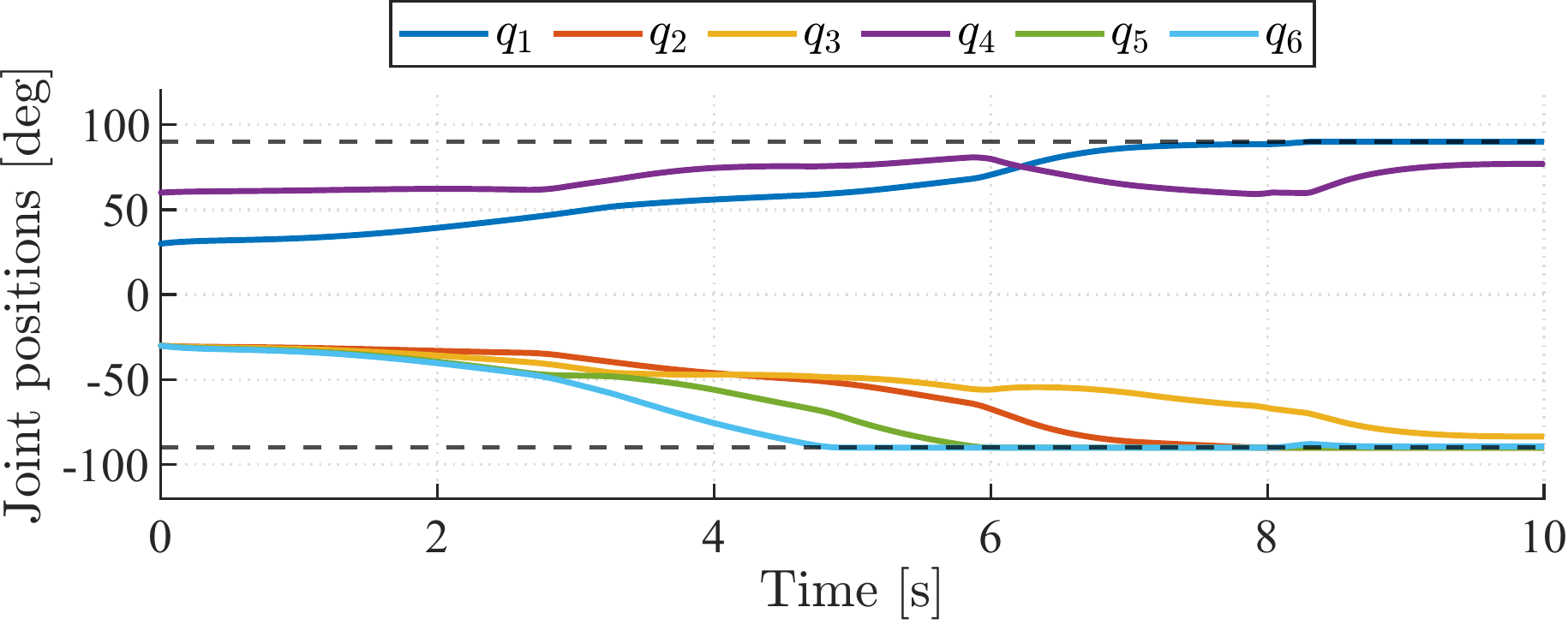}
	\end{subfigure}
	\begin{subfigure}{0.99\columnwidth}
		\centering
		\includegraphics[scale=0.45]{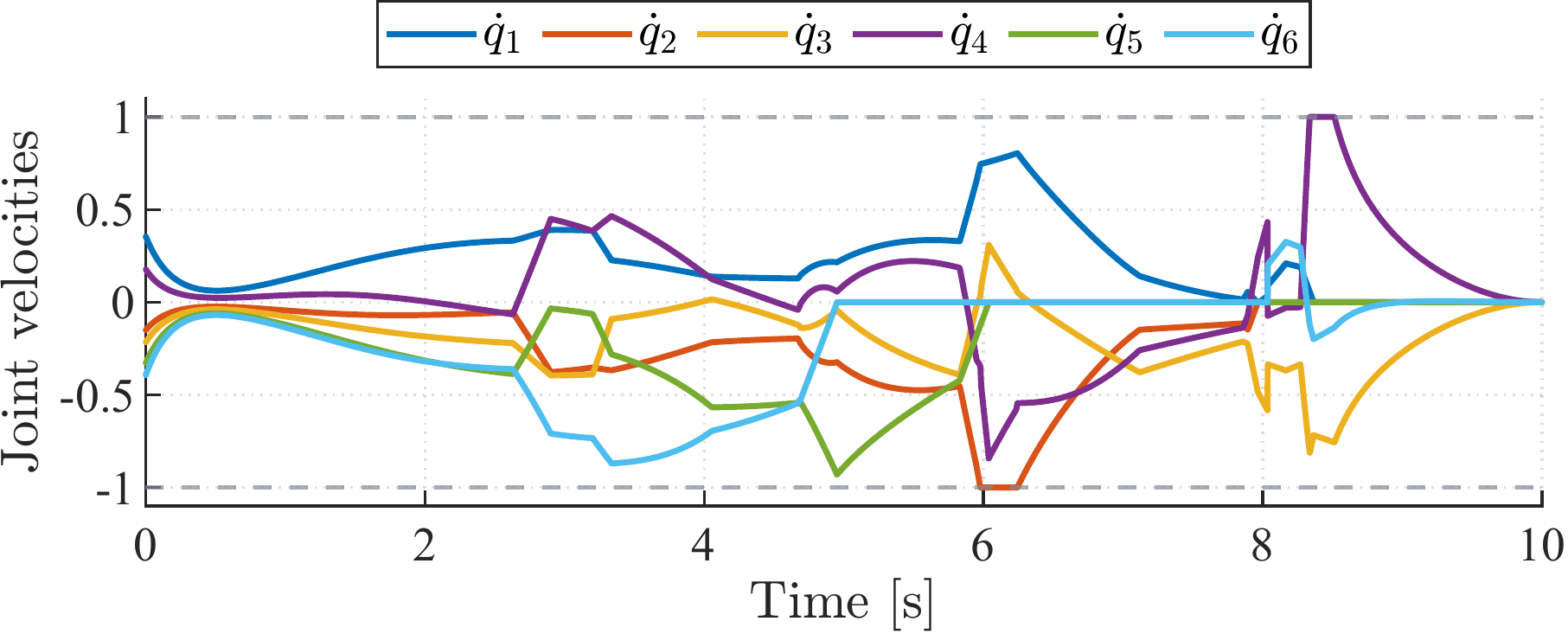} 
	\end{subfigure}
	\caption{Simulation. Evolution of the position and velocity of the joints during task execution. The bounds on the joint motion are indicated by the dashed grey lines.}
	\label{fig:joints}
\end{figure}

\begin{figure}[thpb]		
\centering
	\begin{subfigure}{0.99\columnwidth}
		\centering
		\includegraphics[scale=0.45]{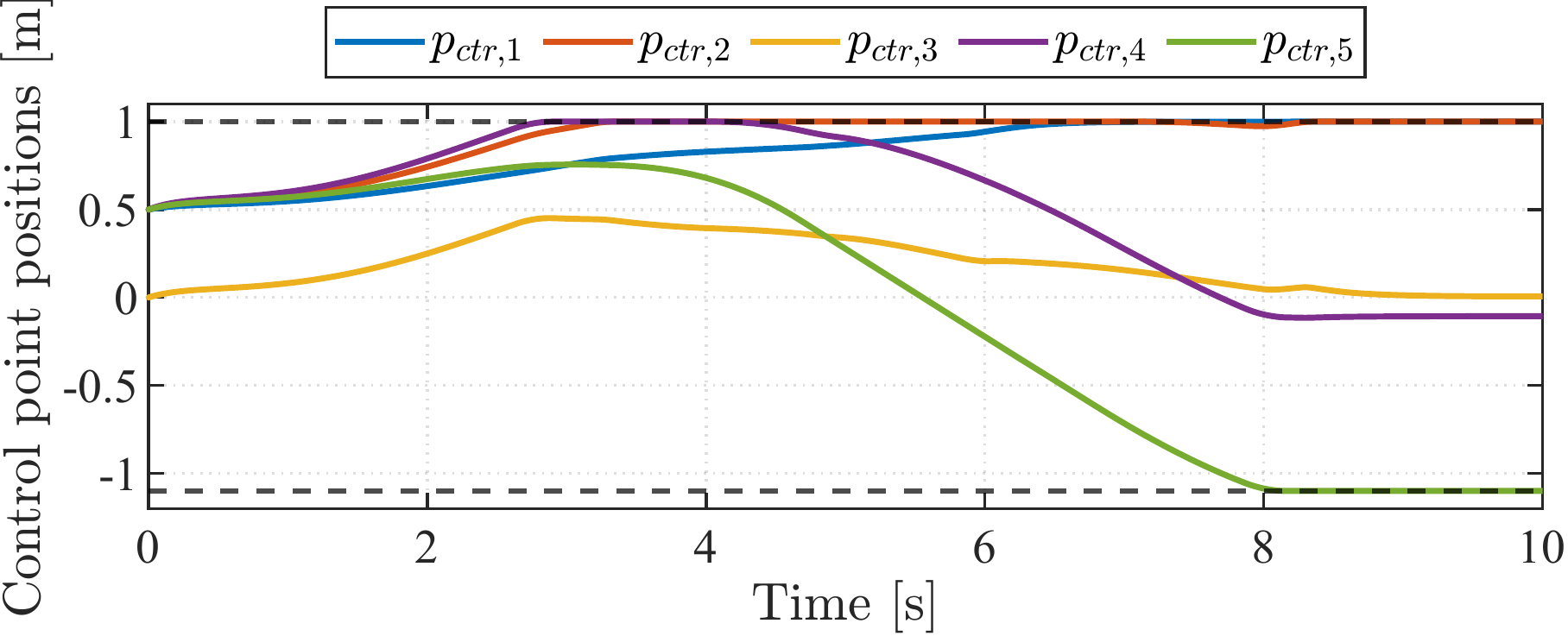}  
		\end{subfigure}	
	\begin{subfigure}{0.99\columnwidth}
		\centering
		\includegraphics[scale=0.45]{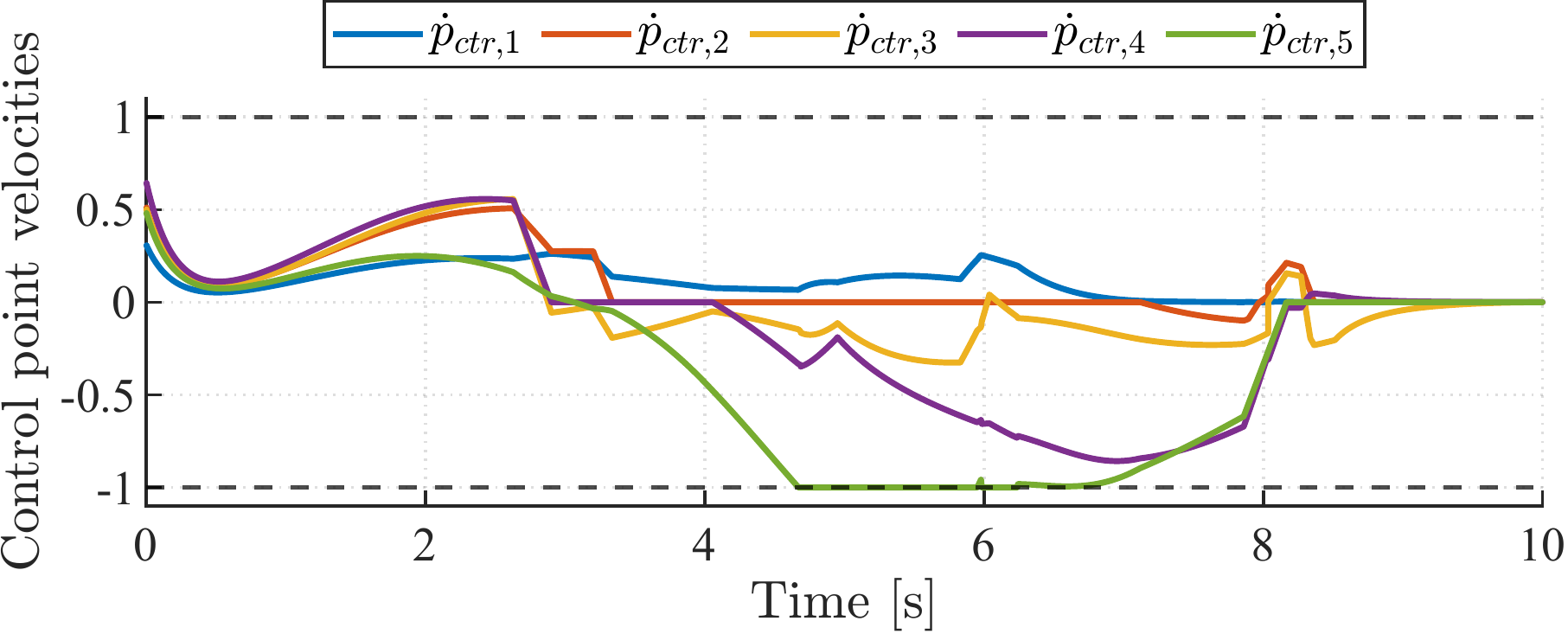}  
		\end{subfigure}
	\caption{Simulation. Evolution of the position and velocity of the control points along the $y$-direction. The Cartesian bounds on the motion of the control points are indicated by the dashed grey lines.}
	\label{fig:cartesian}
\vspace{-10pt}
\end{figure}

The EE begins its motion near to the desired path. As shown in Fig.~\ref{fig:error}, the position error converges quickly and is kept to zero throughout the task execution, except when the task scaling is (mildly) active (i.e., $s^\ast < 1$) to comply with the saturated phases occurring in the joint and Cartesian motion ---see Figs.~\ref{fig:joints} and~\ref{fig:cartesian}. The robot is capable of completing the primary task while satisfying all hard inequality constraints (many of which are saturated).

\subsection{Experiments with the KUKA LWR robot}
The proposed \Cref{basic:vel:single} has been implemented in C++ to perform experimental evaluations with a KUKA LWR IV robot ($n=7$). A position control mode through the KUKA FRI library is used, with sampling time $T = 5$~[ms]. The results of two experiments are presented.

\subsubsection{Cartesian constraints at the elbow}
\begin{figure*}[thpb]
\centering
\begin{subfigure}{.65\columnwidth}
\includegraphics[width=\columnwidth]{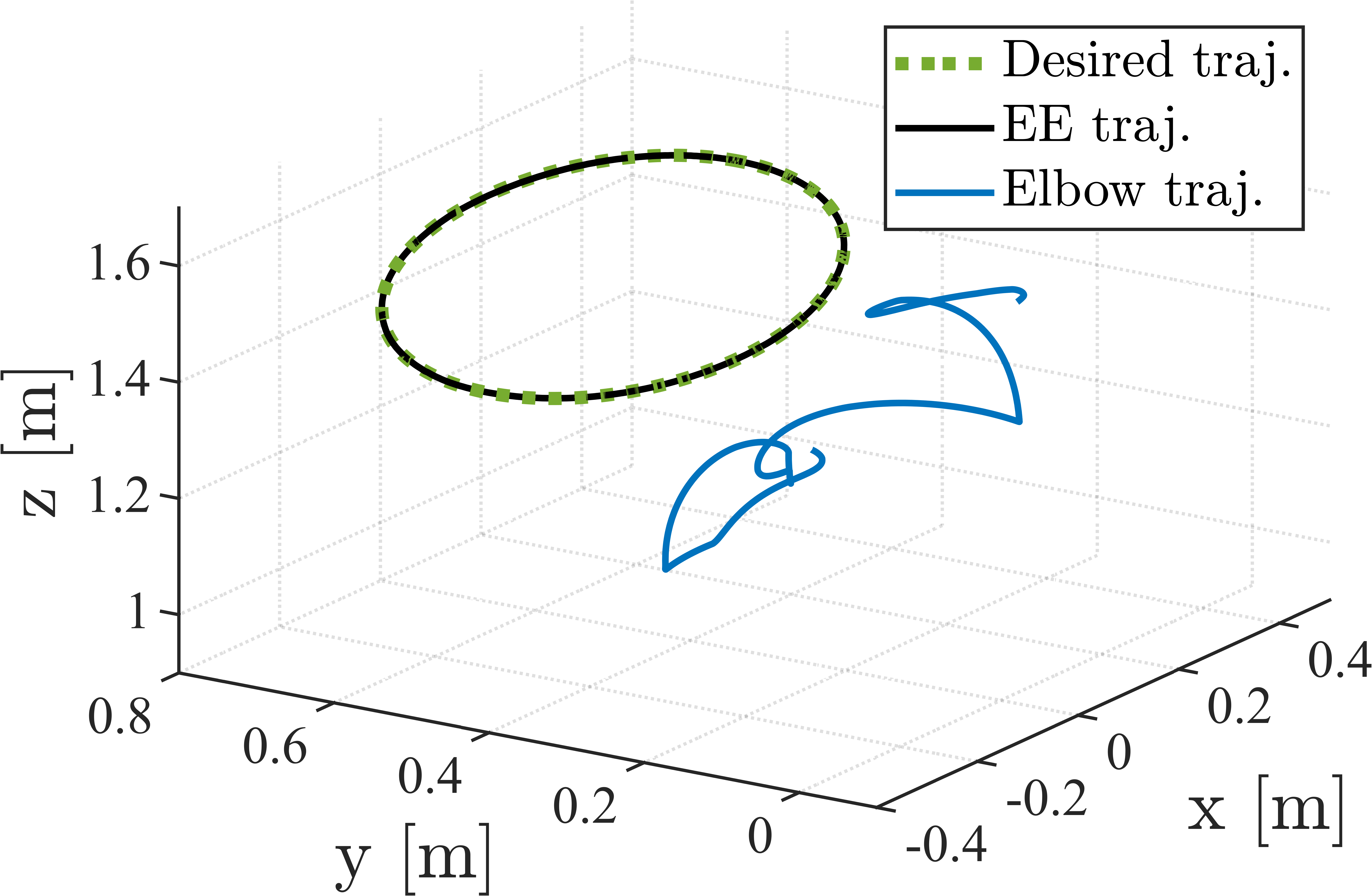}%
\caption{\centering}%
\end{subfigure}\hfill%
\begin{subfigure}{.65\columnwidth}
\includegraphics[width=\columnwidth]{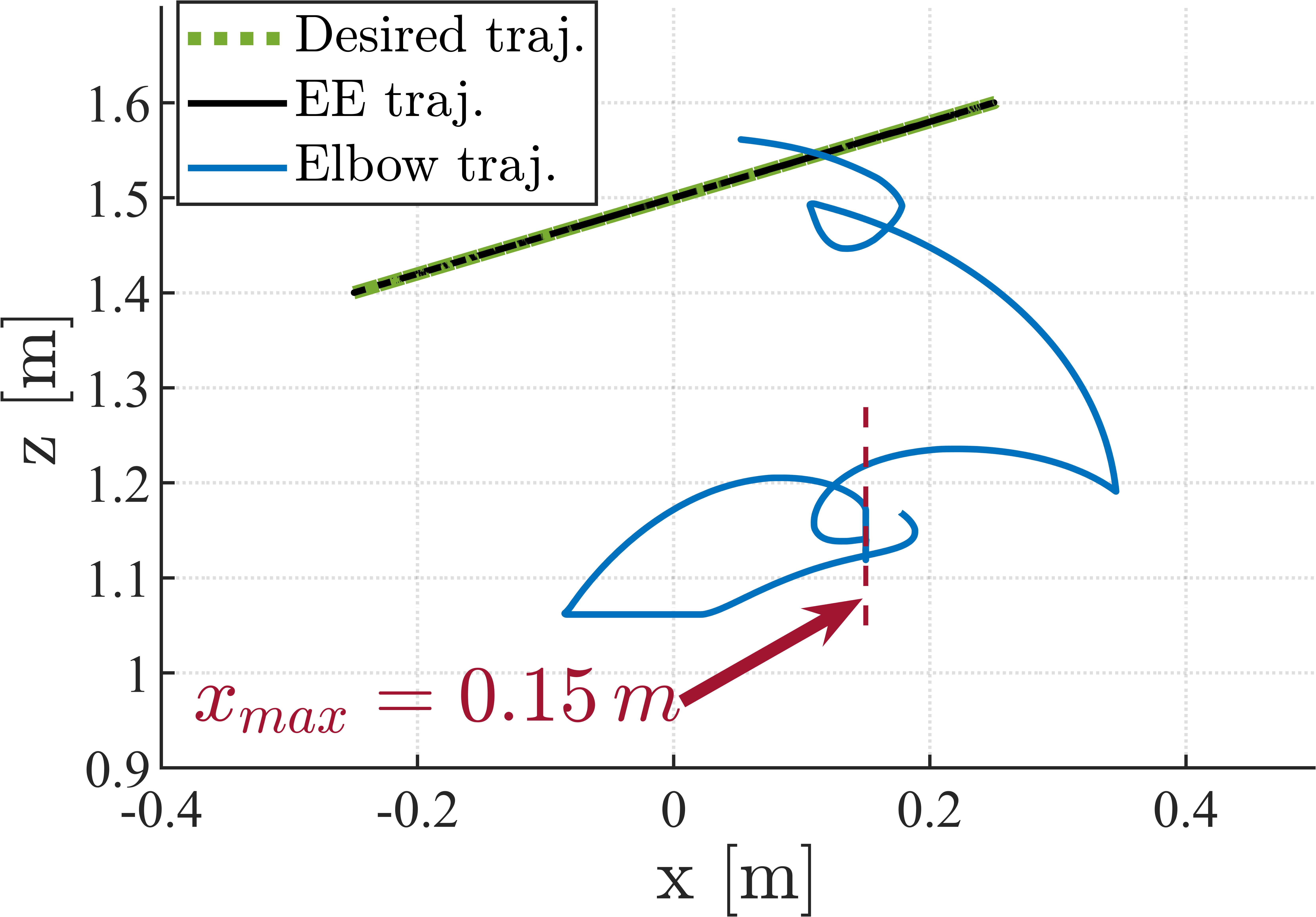}%
\caption{\centering}%
\end{subfigure}\hfill%
\begin{subfigure}{.65\columnwidth}
\includegraphics[width=\columnwidth]{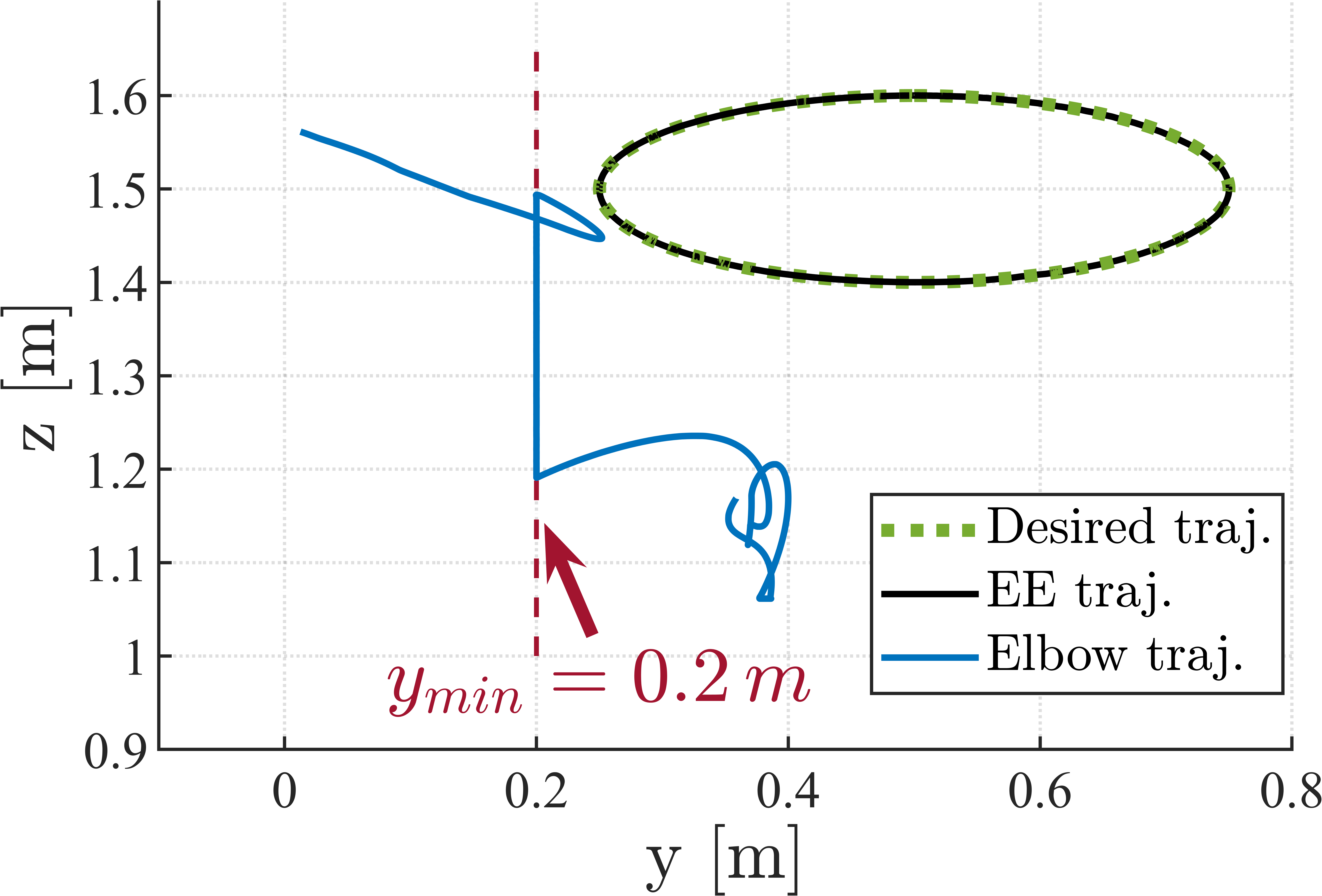}%
\caption{\centering}%
\end{subfigure}\hfill%
\caption{First experiment. (a) The motion executed by the end effector (in black) coincides with the desired circular path (dashed green). The position of the robot elbow (blue traces) satisfies the temporal constraints~(\ref{eq:Car_limits}), both on the $x$-axis (b) and on the $y$-axis (c).}
\label{fig:exp1:view}
\end{figure*}

\begin{figure}[thpb]
\centering
\includegraphics[scale=0.45]{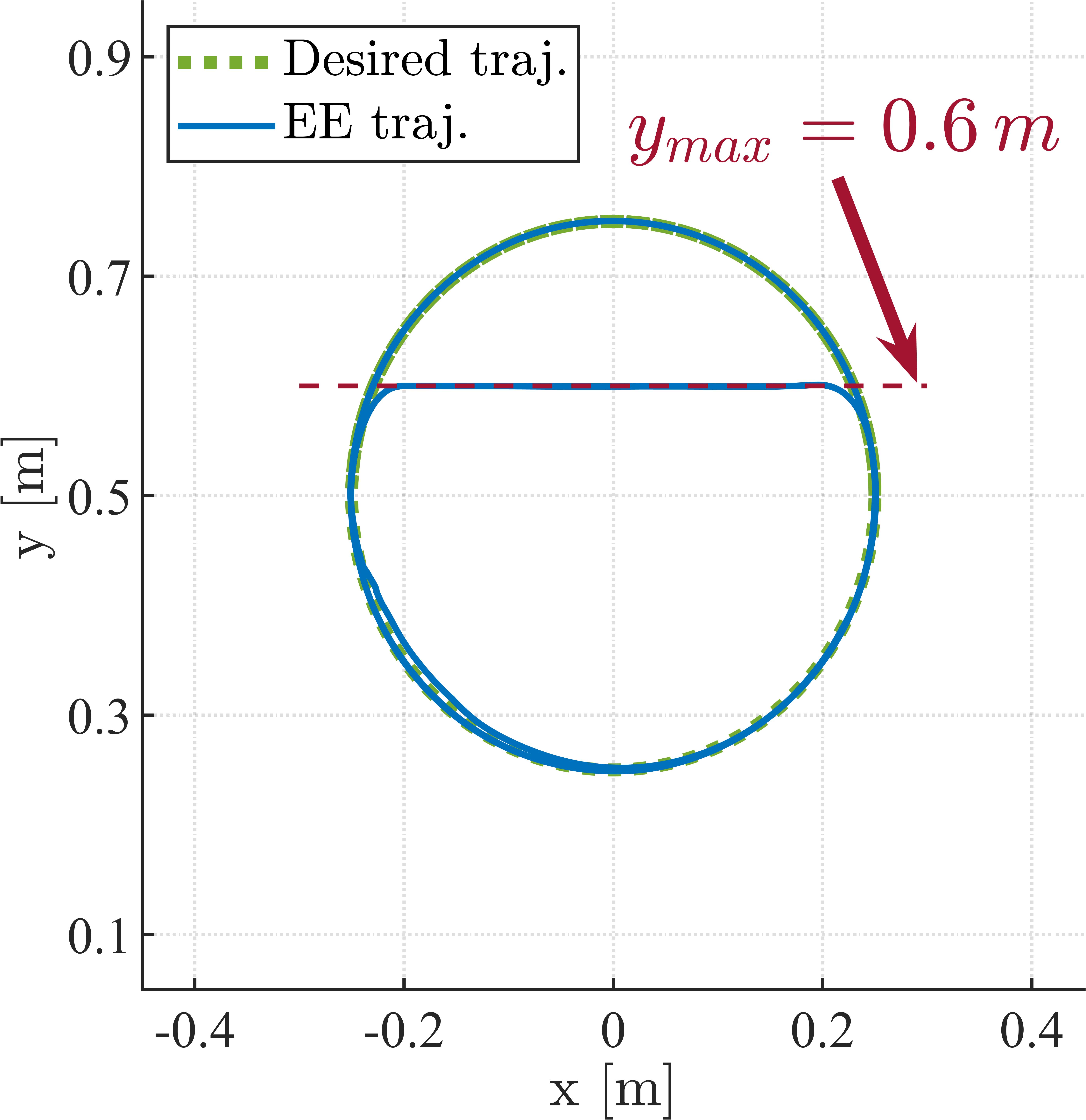}
\caption{Second experiment. The end-effector motion (in blue) coincides with the desired circular path (dashed green) in the first and third rounds, while its position saturates the temporal  constraint~(\ref{eq:Car_limits2}) during the second round.}
\label{fig:exp2:view}
\end{figure}

In the first experiment, the desired EE task is to track three times a 3D circle ($m= 3$), starting on the path with the initial joint configuration
\begin{equation*}
    \qv_0 = (13.50 \ -7.76 \  55.16 \ 79.70 \ 0 \ -6.19 \ 0)^T \ [\deg].
\end{equation*}
Defining the world frame on the lab floor (see, Fig.~\ref{fig:KUKA}), the desired circular path is centered at $\bm{C}=(0 \ 0.5 \ 1.5)^T$~[m], with a radius of $0.25$~[m]. The timing law on the path has a trapezoidal velocity profile, with maximum acceleration $\Ddot{\sigma}=0.15$~[m/s$^2$] and cruise velocity $\dot{\sigma}=0.15$~[m/s]. The control gain matrix is set to $\Km_p =\text{diag}(30,30,30)$. The robot joint limits are set to

\begin{align}
\bm{Q}^{max} &= - \bm{Q}^{min}= \left(170\ 105\ 170\ 120\ 170\ 85\ 170\right)^T \ [\deg],\nonumber\\
\bm{V}^{max} &= - \bm{V}^{min}= \left(20\ 22\ 20\ 26\ 26\ 36\ 36\right)^T \ [\si{\deg \per \second]},\nonumber
\\
\Lambdam^{max} &= - \Lambdam^{min}= \left(30\ 30\ 30 \ 30\ 30\ 30\ 30\right)^T \ [\si{\deg \per \square \second}].\nonumber
\end{align}
A single control point of dimension $d_1=2$ is considered at the robot elbow (joint 4), which has to satisfy the temporal constraints
\begin{equation}
\begin{array}{cc}
\pv_{cp_x,1} \leq 0.15  \ [\si{\meter}], & 16 \leq t \leq 22 \ [\si{\second}], \\[6pt]
\pv_{cp_y,1} \leq 0.2  \ [\si{\meter}],\ \ & 5 \leq t \leq 10 \ [\si{\second}],
\end{array}
\label{eq:Car_limits}
\end{equation}
and the permanent constraints 
\begin{equation}
\begin{array}{cc}
-0.1 \leq \dpv_{cp_x,1} \leq 0.1, & -0.1 \leq \dpv_{cp_y,1} \leq 0.1 \ [\si{\meter/\second}], \\[6pt]
-0.5 \leq \ddpv_{cp_x,1} \leq 0.5, & -0.5 \leq \ddpv_{cp_y,1} \leq 0.5  \ [\si{\meter/\square\second}].
\end{array}
\end{equation}

\begin{figure}[htbp]
\centering
\begin{subfigure}{\columnwidth}
\includegraphics[scale=0.4]{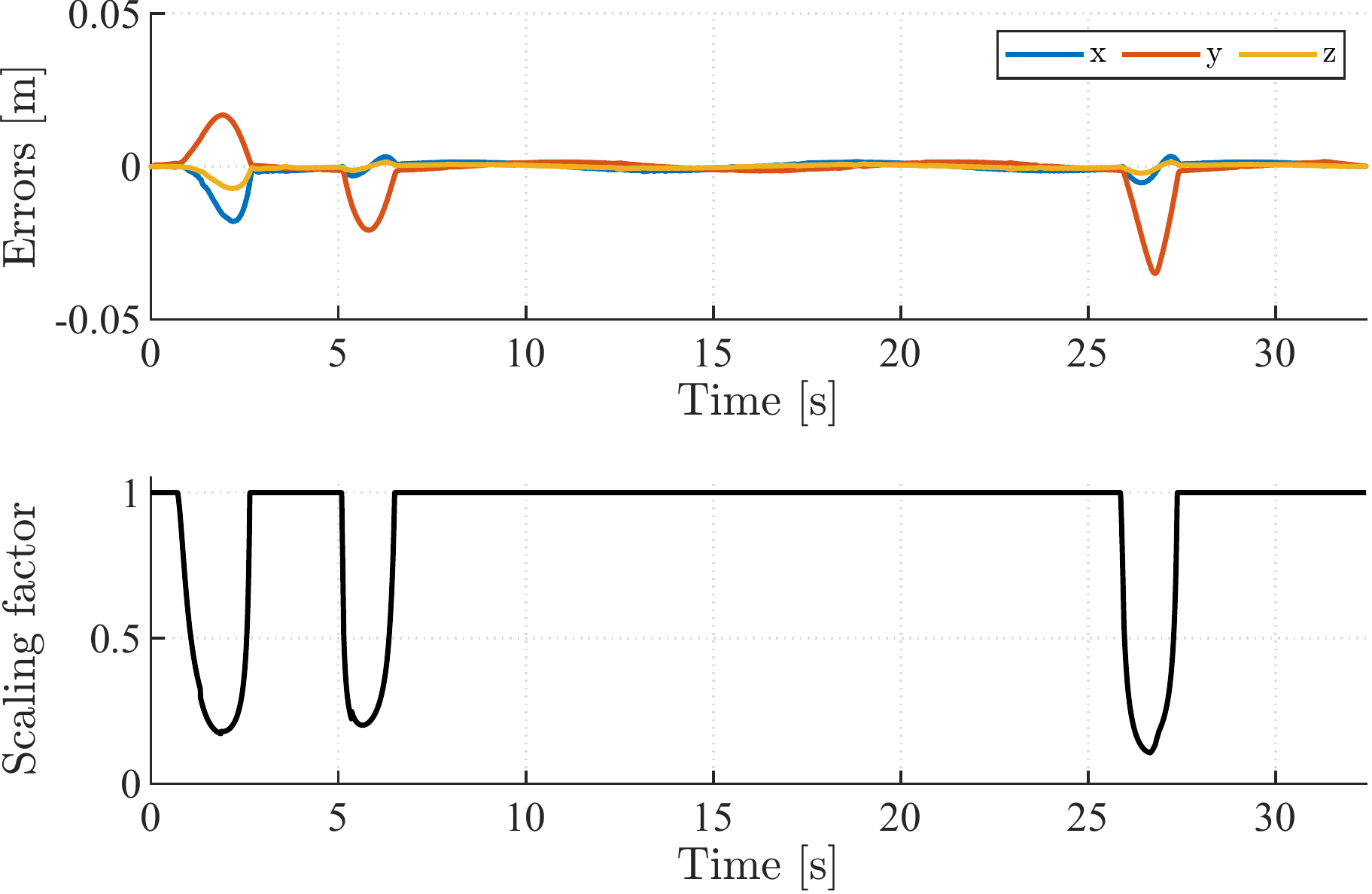}
\end{subfigure}
\\[3pt]
(a)\\[3pt]
\begin{subfigure}{\columnwidth}
\includegraphics[scale=0.4]{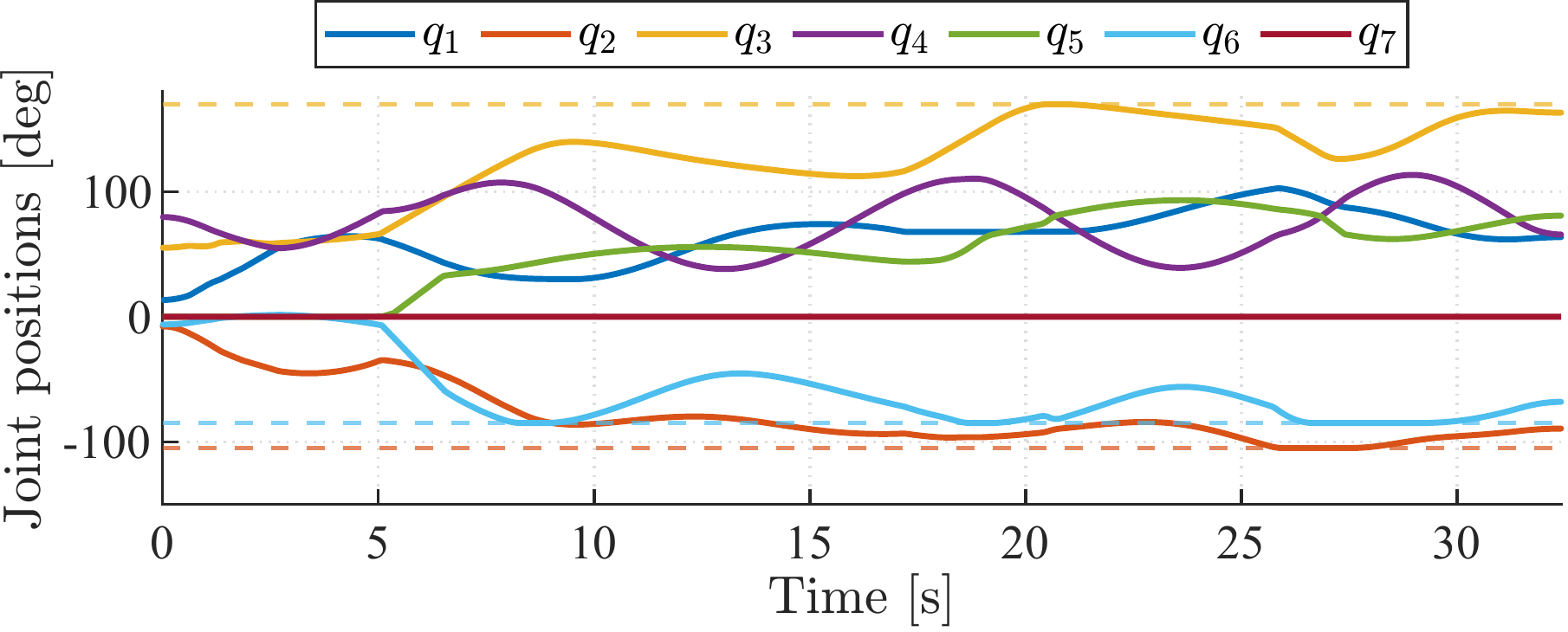}
\end{subfigure}
\begin{subfigure}{\columnwidth}
\includegraphics[scale=0.4]{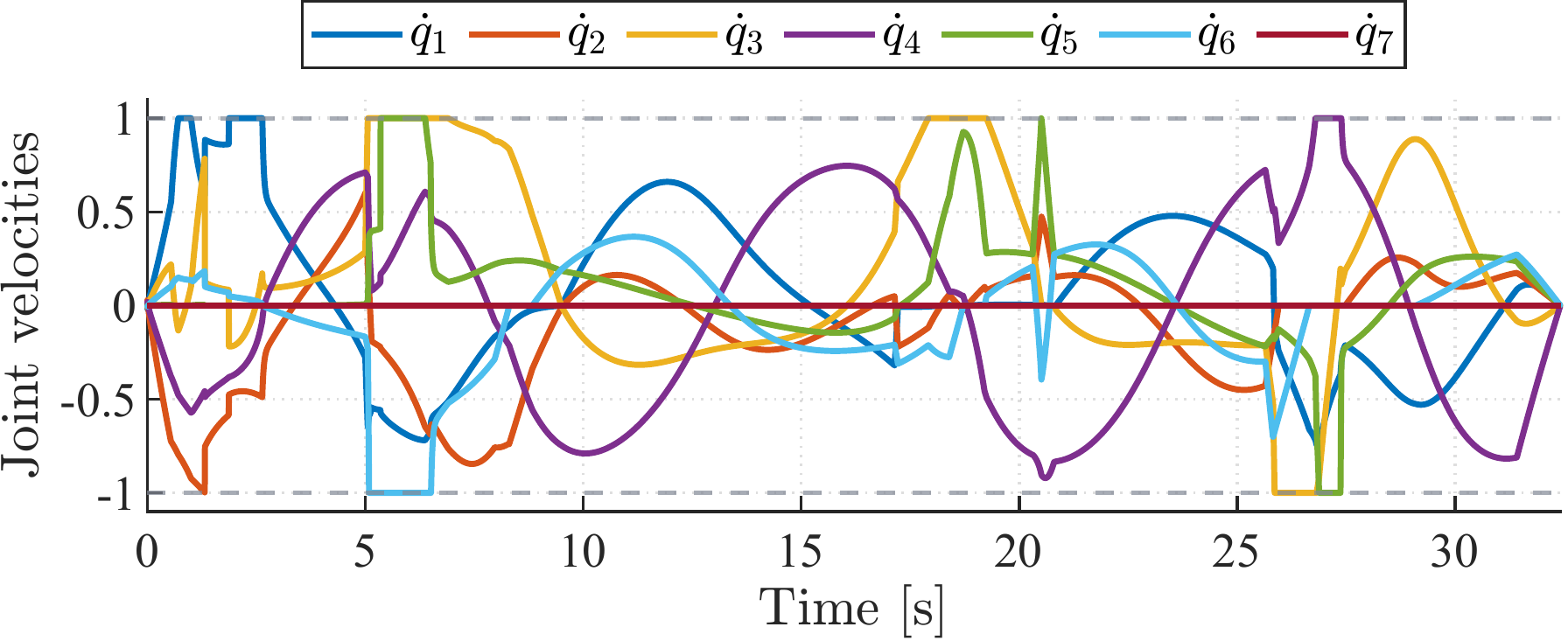}
\end{subfigure}
\\[3pt]
(b)\\[3pt]
\begin{subfigure}{\columnwidth}
\includegraphics[scale=0.4]{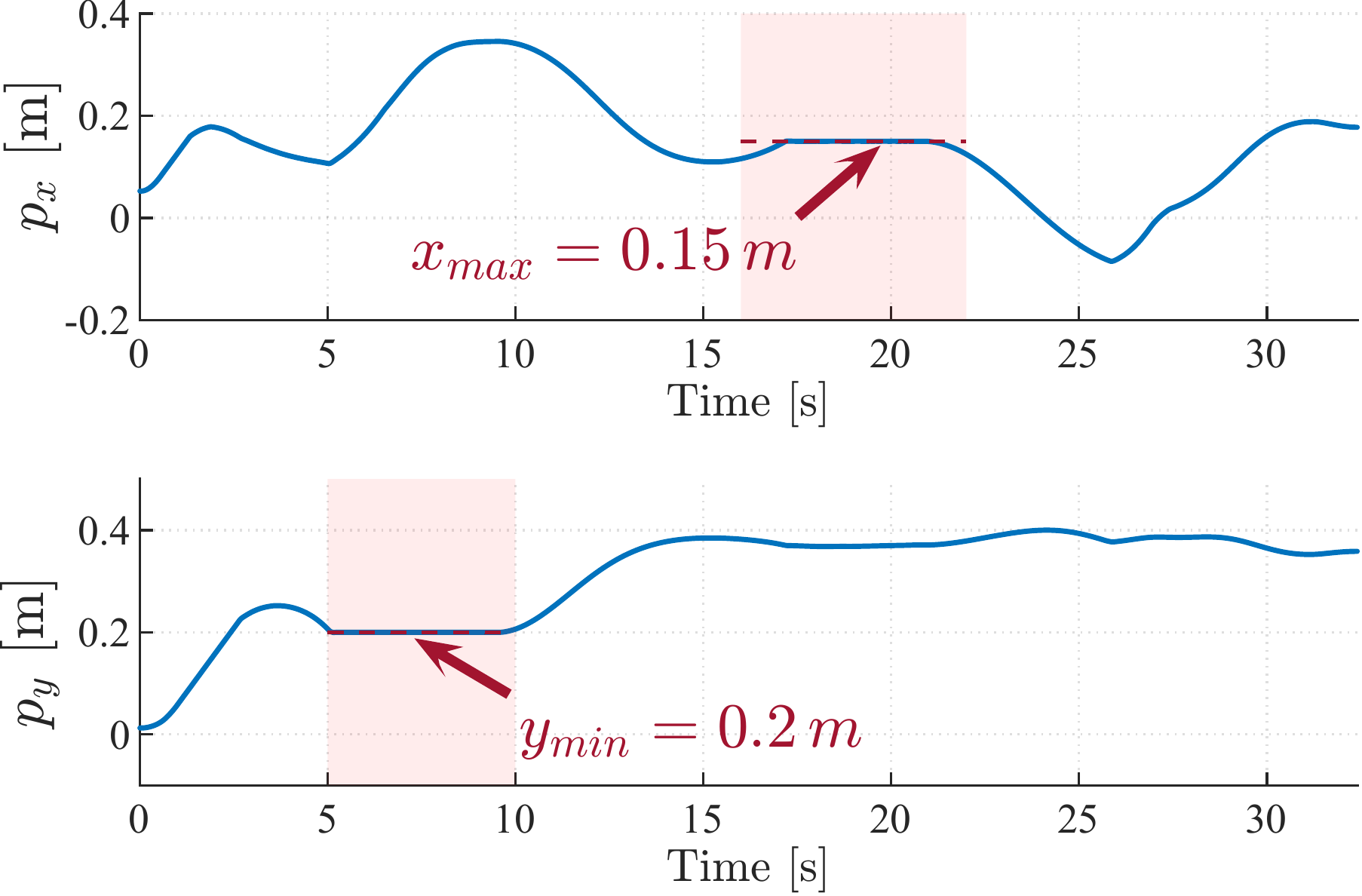}
\end{subfigure}
\begin{subfigure}{\columnwidth}
\ \ \includegraphics[scale=0.4]{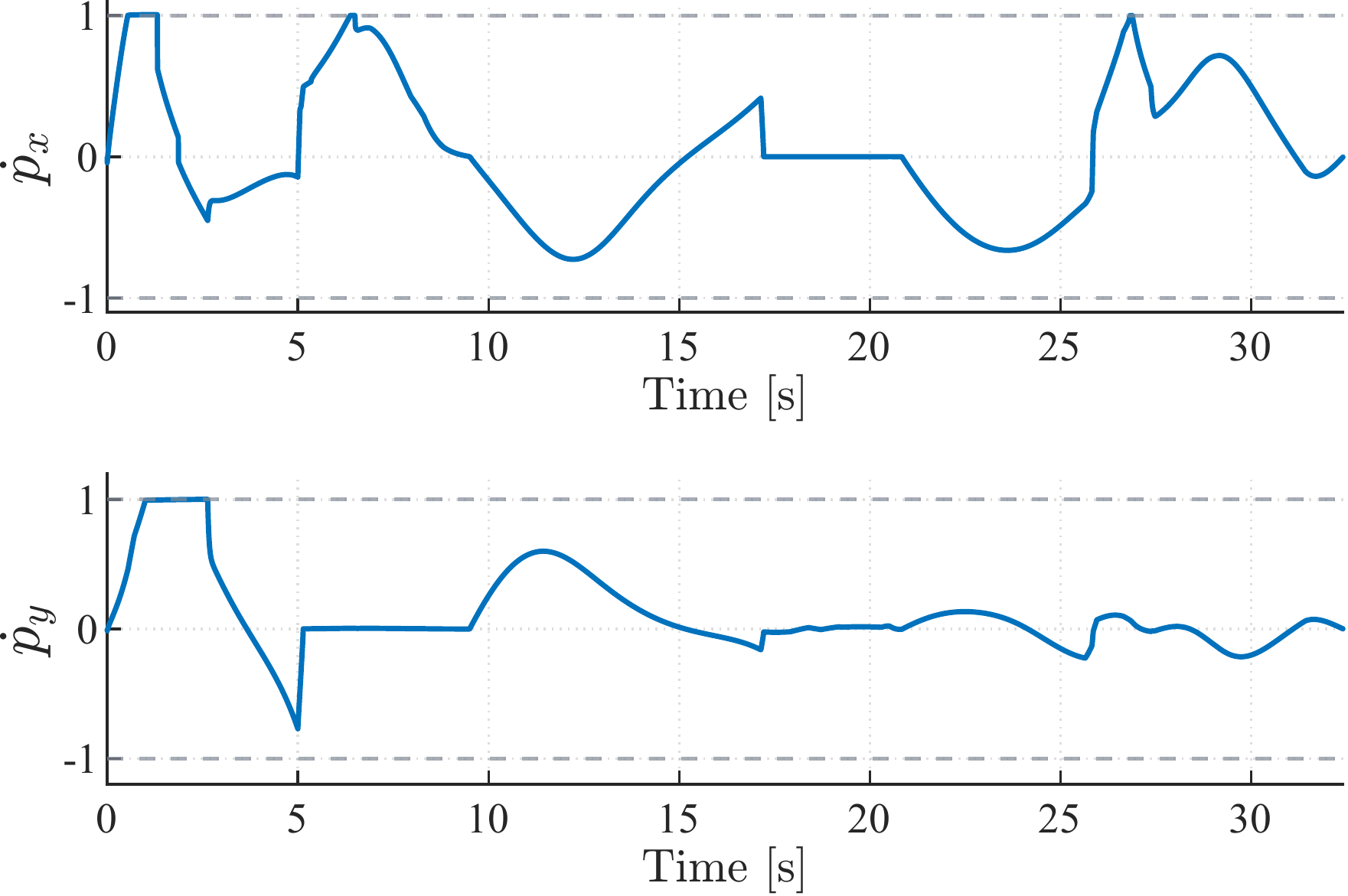}
\end{subfigure}
\\[3pt]
(c)
\caption{First experiment. (a) Error components on the primary task and the related optimal scaling factor. (b) Joint positions and normalized velocities. (c)  The elbow (control point) position and normalized velocity components. The shadowed areas in pink represent the activation period of the constraints~(\ref{eq:Car_limits}). Dashed lines indicate the associated bounds.}
\label{fig:exp1}
\end{figure}

Figure~\ref{fig:exp1:view} shows how the robot executes the desired task by complying with the Cartesian constraints. In Fig.~\ref{fig:exp1}(a), the errors on the primary task are  zero, except when no feasible solution exists under the considered hard constraints. In this case, the robot task is scaled down, i.e., $s^\ast < 1$, while keeping the EE velocity direction tangent to the desired path. In fact, the EE motion in Fig.~\ref{fig:exp1:view} keeps nicely the geometry of the original path. 

The corresponding evolution of the joints in Fig.~\ref{fig:exp1}(b) satisfies the hard joint limits at all times. The frequent saturation in position of joints 2, 3 and 6, as well as of all joint velocities (except for joint 7) clearly illustrates how \Cref{basic:vel:single} exploits the available joint motion capabilities. The motion of the robot elbow (control point) is shown in Fig.~\ref{fig:exp1}(c). When the inequality constraints~(\ref{eq:Car_limits}) are activated/deactivated  (i.e., the shadowed areas), the elbow reacts properly and saturates, if necessary, to stay in the limits. 

\subsubsection{Cartesian constraints at the end-effector}
In the second experiment, a different single control point of dimension $d_1=2$ is chosen, placed coincident with the robot end effector. The primary task of dimension $m = 3$ requires the EE to follow three times a circular path in the $xy$ plane, with a center and radius similar to the previous case. The initial joint configuration is chosen as
\begin{equation*}
   \qv_0 = (13.19\ -8.59\ 54.78\ 89.04\ -0.069\ -10.09\ 0 )^T \ [\deg],
\end{equation*}
with the robot starting on the desired path. In this second case, we raise the maximum velocity and acceleration along the EE trajectory to $\dot{\sigma} = 0.65\,\mbox{[m/s]}$ and $\Ddot{\sigma} = 0.65\,\mbox{[m/s$^2$]}$, respectively. This is to push the robot closer to its physical limits. Accordingly, we use the (symmetric) joint limits provided by the manufacturer
\begin{align}
\bm{Q}^{max} &= \left(170\ 120\ 170\ 120\ 170\ 120\ 170\right)^T  [\deg],\nonumber \\
\bm{V}^{max} &= \left(100\ 110\ 100\ 130\ 130\ 180\ 180\right)^T  [\si{{\deg}/{\second}]},\nonumber
\end{align}
with $\bm{Q}^{min} \!=\! -\bm{Q}^{max}$ and $\bm{V}^{min} \!=\!- \bm{V}^{max}$.
The joint acceleration limits are set to $\Lambdam^{max}_j \!=\! - \Lambdam^{min}_j=300 \ [\si{\deg \per \square \second}]$, for $j=1,\dots, 7$. The control point should satisfy the temporal constraint
\begin{equation}
\begin{array}{cc}
\pv_{cp_y,1} \leq 0.6  \ [\si{\meter}], & 2.5 \leq t \leq 4.5 \ [\si{\second}],
\end{array}
\label{eq:Car_limits2}
\end{equation}
as well as the permanent constraints 
\begin{equation}
\begin{array}{cc}
-0.7 \leq \dpv_{cp_x,1} \leq 0.7, & -0.7 \leq \dpv_{cp_y,1} \leq 0.7 \ [\si{\meter/\second}], \\[6pt]
-1.5 \leq \ddpv_{cp_x,1} \leq 1.5, & -1.5 \leq \ddpv_{cp_y,1} \leq 1.5  \ [\si{\meter/\square\second}].
\end{array}
\end{equation}

\begin{figure}[htbp]
\centering
\begin{subfigure}{\columnwidth}
\includegraphics[scale=0.4]{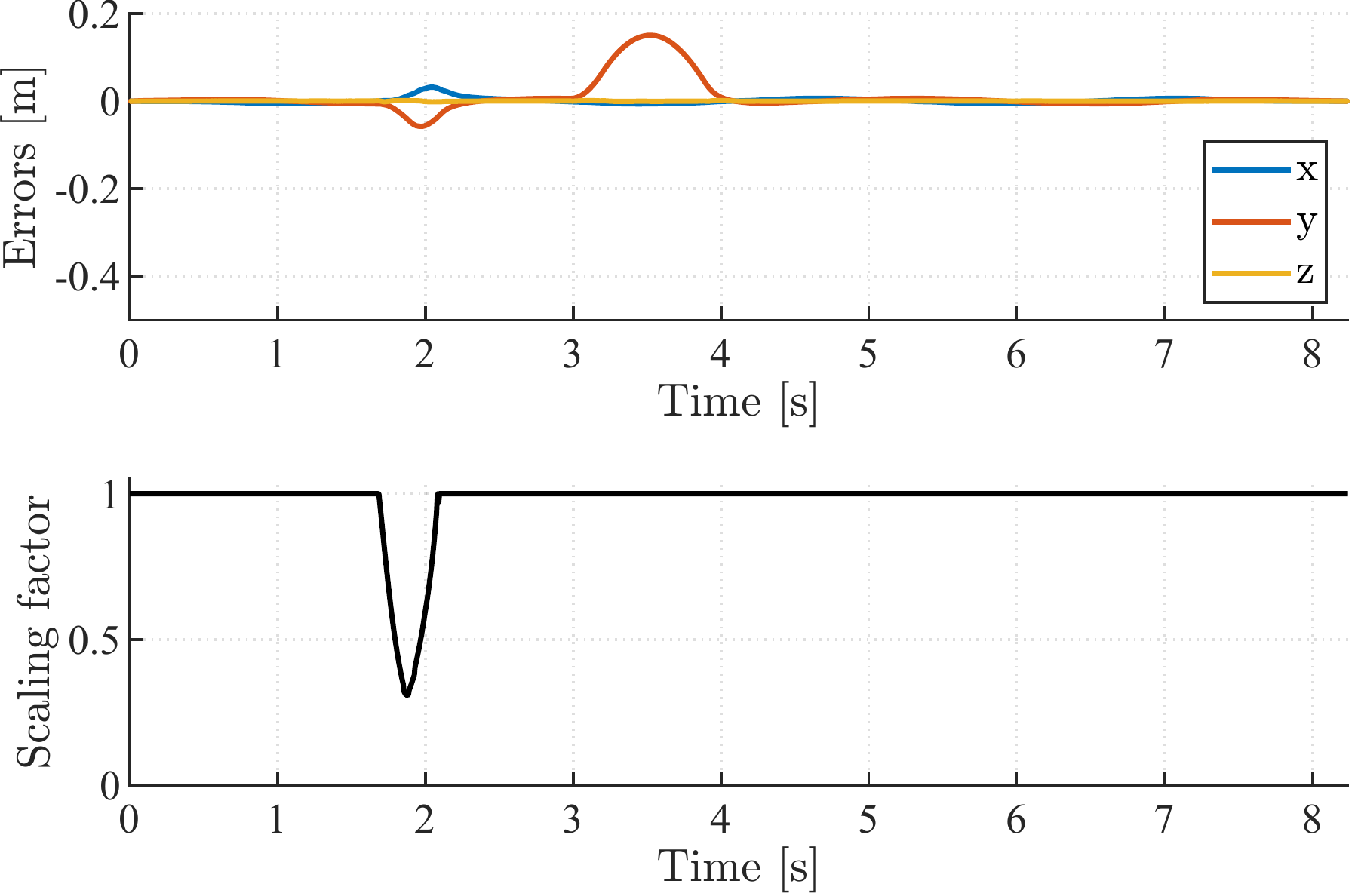}
\end{subfigure}
\\[3pt]
(a)\\[3pt]
\begin{subfigure}{\columnwidth}
\includegraphics[scale=0.4]{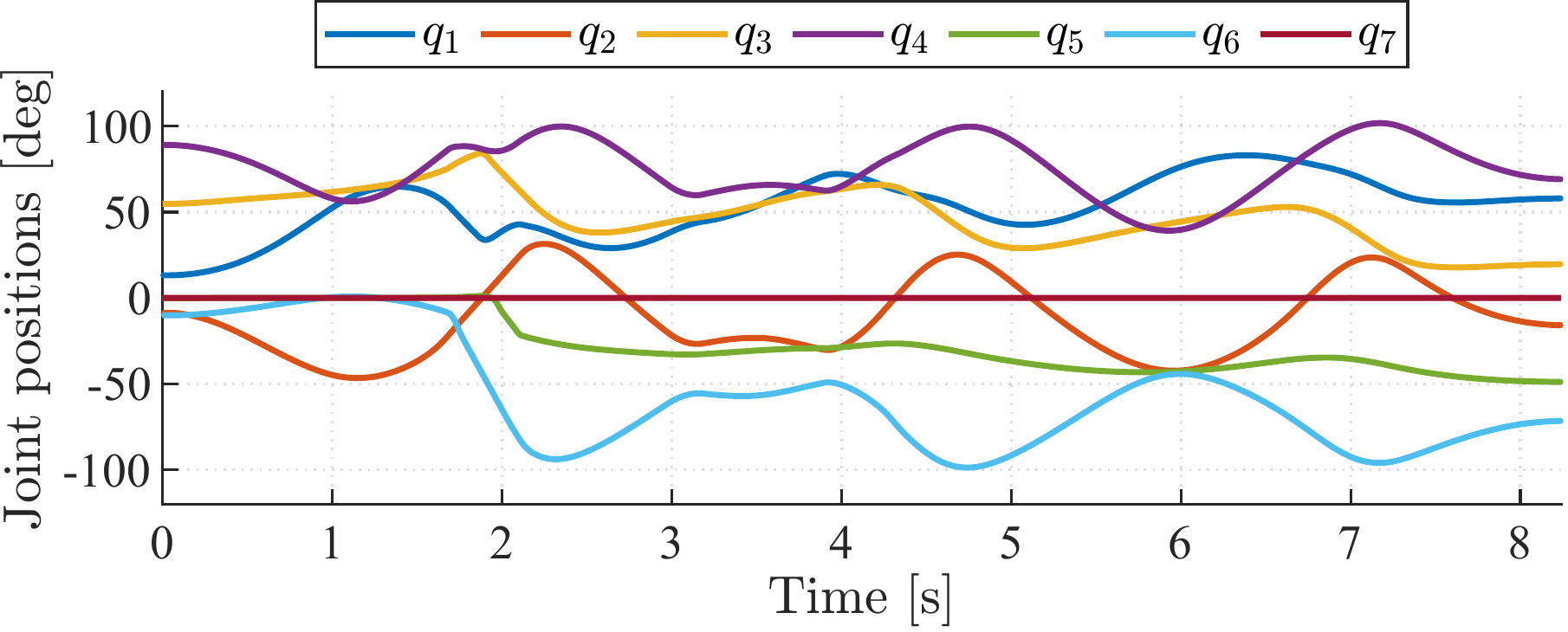}
\end{subfigure}
\begin{subfigure}{\columnwidth}
\includegraphics[scale=0.4]{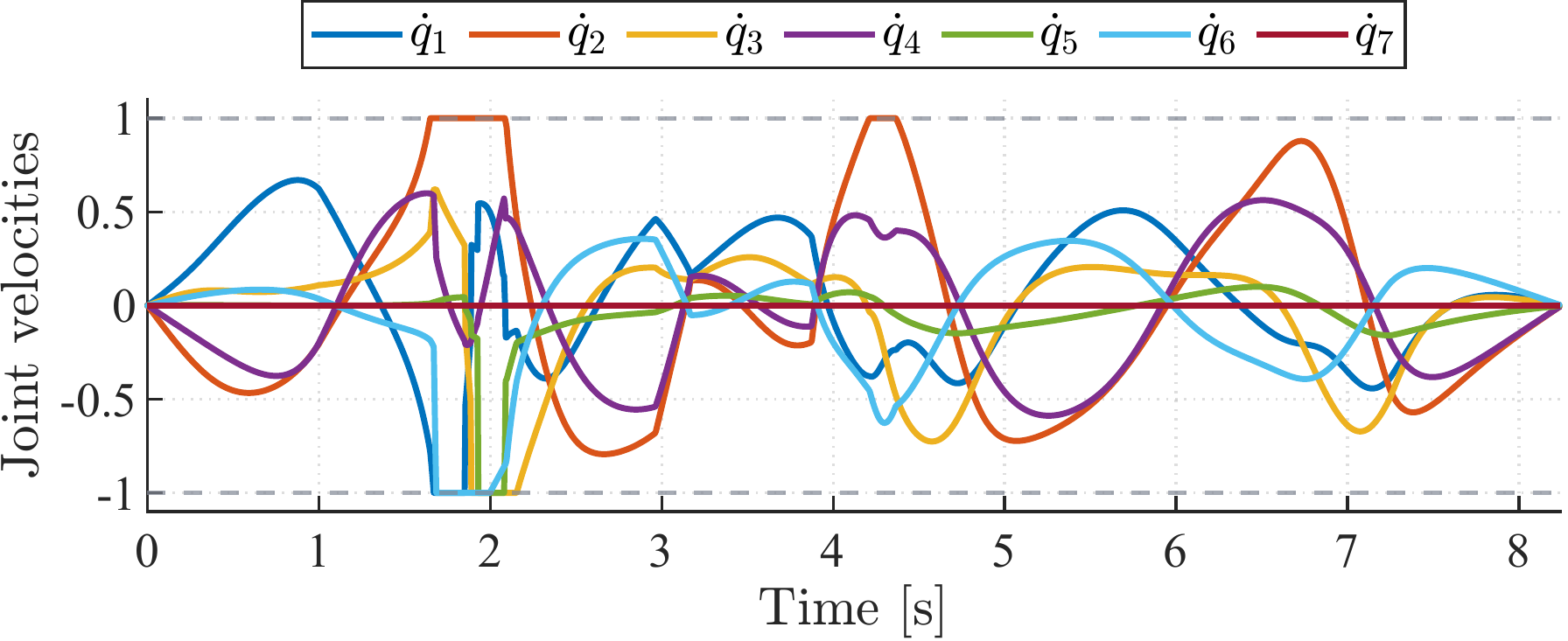}
\end{subfigure}
\\[3pt]
(b)\\[3pt]
\begin{subfigure}{\columnwidth}
\includegraphics[scale=0.4]{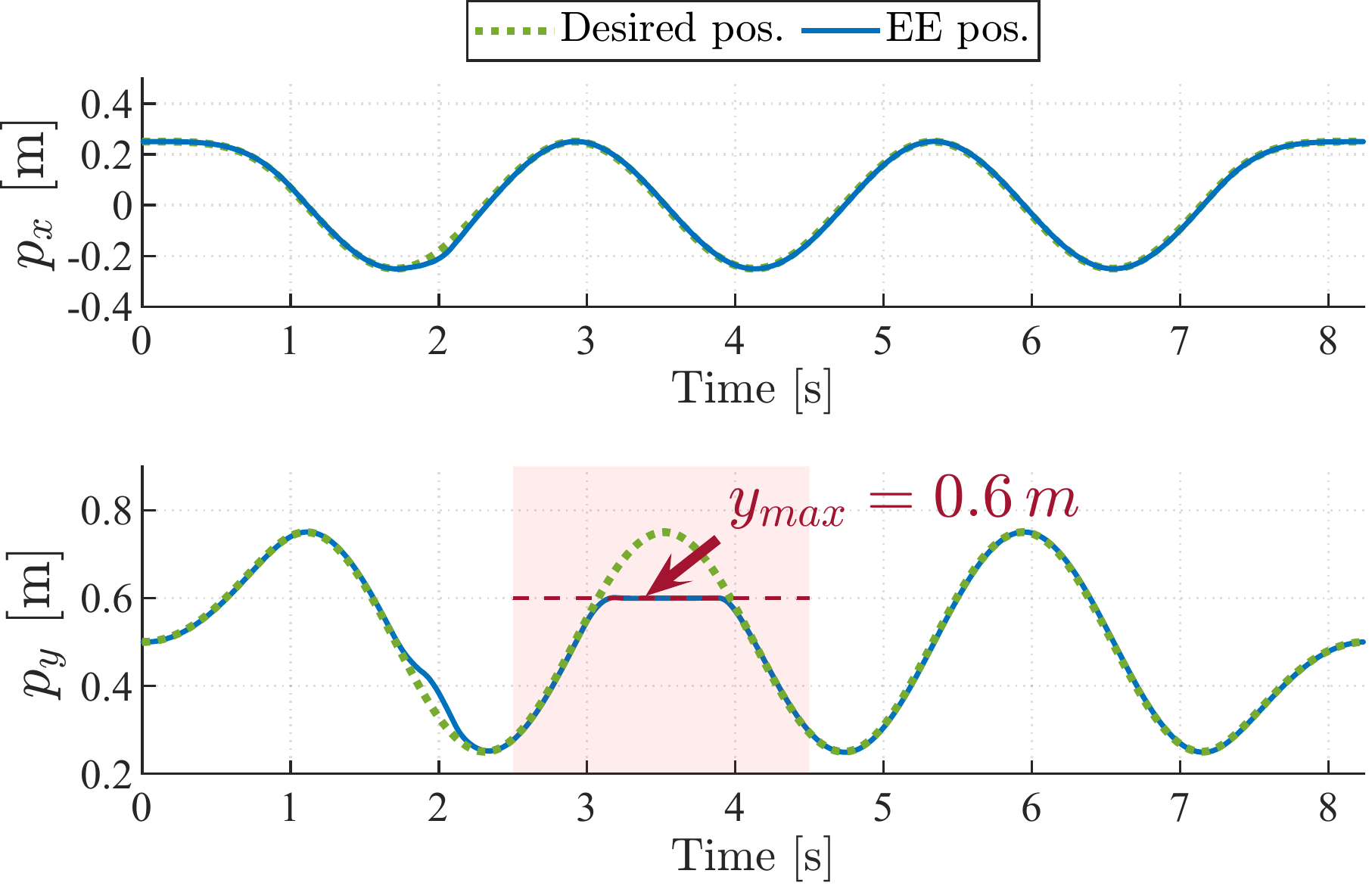}
\end{subfigure}
\begin{subfigure}{\columnwidth}
\ \ \includegraphics[scale=0.4]{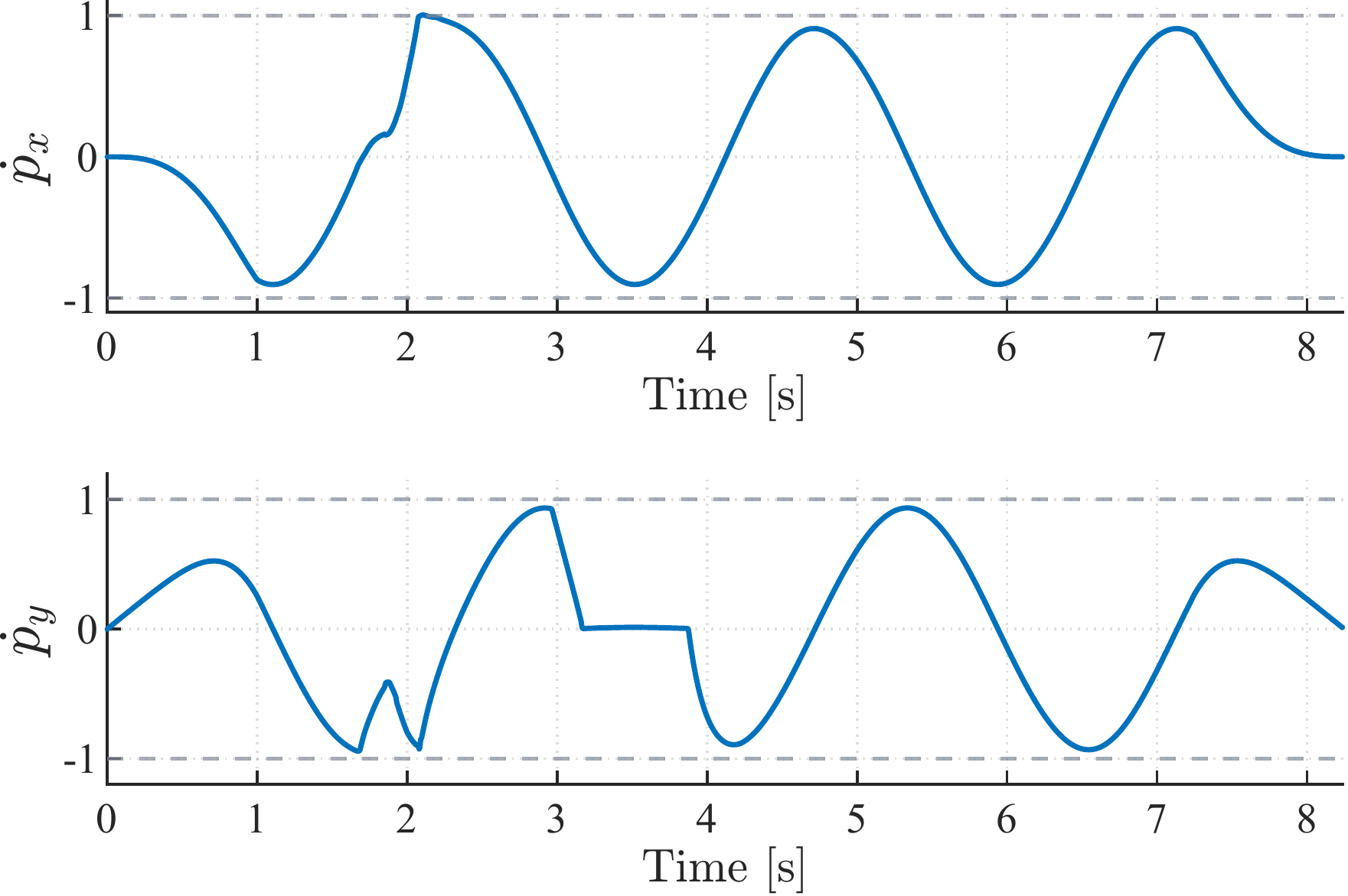}
\end{subfigure}
\\[3pt]
(c)
\caption{Second experiment. (a) Error components on the primary task and the related optimal scaling factor. (b) Joint positions and normalized velocities. (c) The end-effector (control point) position and normalized velocity components. The shadowed area in pink represents the activation period of the constraint~(\ref{eq:Car_limits2}).  Dashed lines indicate the associated bounds.}
\label{fig:exp2}
\end{figure}

Figure~\ref{fig:exp2:view} shows the execution of the task using~\Cref{basic:vel:single}. In Fig.~\ref{fig:exp2}(a), the errors on the EE task increase around $t=2$~[s], where the task scaling factor is applied in order to be able to satisfy the velocity limits of joints 1, 2 and 6 (see Fig.~\ref{fig:exp2}(b)). The EE error along the $y$ direction between $t=3$ and $t=4$~[s] is large, because of the simultaneous activation of the (hard) temporal constraint~(\ref{eq:Car_limits2}), which is in fact inconsistent with the primary task. In this case, there is no use in scaling down the task as done instead in the former event. Moreover, while the EE position saturates at the imposed maximum limit in the $y$-direction, a complete fulfilment of the other task components along the $x$ and $z$ axes is still kept, see Fig.~\ref{fig:exp2}(c). 

\section{Conclusions}
\label{sec:conclusion}
We have presented a generalized null-space saturation algorithm for the kinematic control of redundant robots to realize the primary task under hard inequality constraints in the joint and Cartesian spaces. All hard constraints are equally enforced and the task is automatically scaled in an optimal fashion when no feasible solution exists. The presented case studies have proven the efficiency of the  
approach to handle any possible simultaneous (de-)activation of joint and/or Cartesian inequality constraints without any oscillatory behavior. The ability to handle time-dependent constraints makes the method easy to be integrated in any sensor-based strategy, e.g., for online/dynamic collision avoidance in human-robot collaborative applications~\cite{khatib2021human}. As for the original SNS algorithm~\cite{flacco2015control}, the approach can be developed to include multiple operational tasks with  priorities, keeping the entire set of hard inequality constraints out of the stack of equality tasks. Along similar lines, one can consider to move the commands to the acceleration level, making them suitable for torque-controlled robotic systems.

\bibliographystyle{IEEEtran}
\bibliography{refs}

\end{document}